\documentclass{article}
\usepackage[preprint]{molo_style}
\usepackage{placeins,longtable,multirow,xcolor,colortbl,tabularx}
\usepackage[colorlinks=true,linkcolor=accent,citecolor=accent,urlcolor=accent]{hyperref}
\usepackage[capitalise,nameinlink]{cleveref}
\crefname{section}{Sec.}{Secs.}\Crefname{section}{Sec.}{Secs.}
\crefname{subsection}{Sec.}{Secs.}\Crefname{subsection}{Sec.}{Secs.}
\crefname{subsubsection}{Sec.}{Secs.}\Crefname{subsubsection}{Sec.}{Secs.}
\crefname{appendix}{App.}{Apps.}\Crefname{appendix}{App.}{Apps.}
\crefname{subappendix}{App.}{Apps.}\Crefname{subappendix}{App.}{Apps.}
\crefname{figure}{Fig.}{Figs.}\Crefname{figure}{Fig.}{Figs.}
\crefname{table}{Tab.}{Tabs.}\Crefname{table}{Tab.}{Tabs.}
\crefname{equation}{Eq.}{Eqs.}\Crefname{equation}{Eq.}{Eqs.}
\crefname{algorithm}{Alg.}{Algs.}\Crefname{algorithm}{Alg.}{Algs.}

\hypersetup{pdfauthor={Pedro M. P. Curvo, Mohammad Mahdi Derakhshani, Gertjan J. Burghouts, Jan-Willem van de Meent, Cees G. M. Snoek},pdftitle={Mixture of Looped Modalities: Untying Depth Across Modalities}}
\title{Mixture of Looped Modalities\\[3pt]
\large Untying Depth Across Modalities}
\author{Pedro M.\ P.\ Curvo$^{1,4,*}$\quad Mohammad Mahdi Derakhshani$^{1,4,*}$\quad Gertjan J.\ Burghouts$^{3,\dagger}$\\
\bfseries Jan-Willem van de Meent$^{2,4,\dagger}$\quad Cees G.\ M.\ Snoek$^{1,4,\dagger}$\\
\normalfont $^{1}$VISLab\quad $^{2}$AMLab\quad $^{3}$TNO, Intelligent Imaging\quad $^{4}$University of Amsterdam\\
\normalfont\texttt{\{p.m.pombeirocurvo,\,m.m.derakhshani\}@uva.nl}\\[2pt]
\normalfont\small $^{*}$Equal co-authorship, in random order.\quad $^{\dagger}$Equal supervision.}
\begin{document}
\raggedbottom
\setlength{\textfloatsep}{12pt plus 2pt minus 2pt}
\setlength{\intextsep}{10pt plus 2pt minus 2pt}
\maketitle
\vspace{-22pt}
\begin{abstract}
Multimodal transformers usually pass every token through the same blocks the same number of times. Mixture-of-Transformers relaxes this by giving each modality its own weights, yet all modalities still share one depth. Looped transformers make that depth adjustable at inference by reusing their blocks, but a looped multimodal core still advances every modality together. Combining the two, we introduce \textbf{Mixture of Looped Modalities (MoLo)}, a generative transformer that unties depth across modalities by reusing each modality's core weights on its own depth grid, while joint attention lets an advancing modality read the others' latest state. Because training samples the grids independently and supervises every pass of the generated modality toward its own target, one checkpoint runs any number of loops per modality without retraining. We train one model with 188M trainable parameters for text-to-image and image-to-text generation with continuous flows and ask where depth helps. In both directions, the generated modality improves with more loops, whereas the conditioning modality needs at most one. On five prompt-following benchmarks, MoLo, which also performs image-to-text generation, has a higher average than \textbf{13 of 15} published text-to-image generators, including all four above 6B parameters, while using \textbf{under 3\% of their parameters}.
\end{abstract}

\begingroup
\setlength{\intextsep}{6pt}
\begin{figure}[H]
\setlength{\abovecaptionskip}{3pt}
\centering
\includegraphics[width=\linewidth]{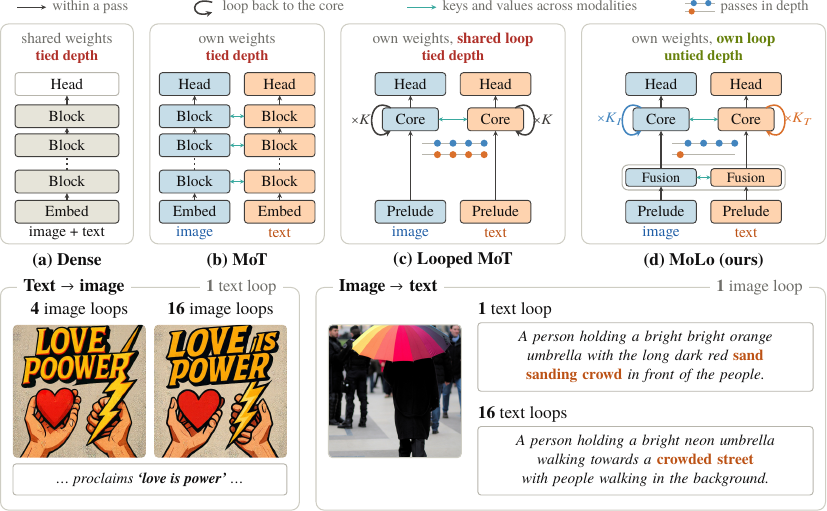}
\caption{\textbf{MoLo unties depth across modalities.} (a)~A dense transformer shares weights and depth. (b)~MoT gives each modality its own weights, (c)~looping its core makes depth adjustable but ties it across modalities, and (d)~MoLo gives each modality its own loop. Bottom: with input and noise fixed, more loops of the generated modality fix a poster's lettering and what a caption describes.}
\label{fig:cover}
\end{figure}
\endgroup

\clearpage
\section{Introduction}
\label{sec:introduction}

Transformers~\citep{vaswani2017attention} have become a standard backbone for
generative modeling, with increases in model size driving many advances in
image and video generation as well as scientific
applications~\citep{peebles2023,esser2024sd3,polyak2024moviegen,nvidia2025cosmos,hayes2025esm3}.
As these models grow, how they spend their computation becomes a central
design choice, especially in models that read and generate several
modalities, interleaving their tokens in one sequence or conditioning one on
another~\citep{chameleon2024,zhou2024transfusion,xie2024showo,bao2023unidiffuser}.
Such a model must also divide its computation among modalities, whose needs
can differ. In most of these networks, however, the division is fixed by the
architecture: every token, whatever its modality, is processed by the same blocks, the same number of times.

This uniformity takes two forms: 1) Because every modality passes through the
same blocks, multimodal transformers are dense in their parameters: all tokens
share the same projections and feed-forward layers, despite the different
statistics of their modalities; 2) Because every modality passes through these
blocks the same number of times, they are also dense in their depth, which is
fixed once the network is trained. In both cases, how the network treats one
modality is tied to how it treats all others, through shared weights in the
first form and shared computation in the second form.

The first form of density has already been relaxed. MMDiT
blocks~\citep{esser2024sd3} and Mixture-of-Transformers
(MoT)~\citep{liang2024mot} untie parameters across modalities: each token is
processed by the weights of its own modality, while self-attention remains
global over the multimodal sequence. The modalities therefore keep
interacting in every layer, and MoT matches dense multimodal
transformers at a fraction of their pretraining compute. Depth, however, is
still shared. A MoT block updates every modality at every layer, so a modality
that only conditions the output, or that has far fewer dimensions, passes
through as many blocks as the one being generated.

Looped transformers make depth a runtime
choice~\citep{dehghani2019universal,geiping2025recurrent}. They reuse the
same blocks across iterations, so a trained model can run at different depths
with a fixed parameter count. In flow-based
generation~\citep{lipman2023,liu2023rectified,albergo2023interpolants},
\citet{derakhshani2026lift} propose the Loop Flow Transformer (LiFT), which
targets each loop at a linear interpolation between the model's initial
estimate and the flow-matching target, and report that it keeps improving with
more loops at inference than in training. LiFT generates a single modality, whereas
Looped-DiT~\citep{chng2026loopeddit} applies looping to text-to-image
generation with a shared MMDiT core. Hence, in Looped-DiT a single loop
count governs both modalities, so every additional pass increases image and
text depth \textit{together}. Making depth adjustable per modality requires
loop counts that can be set independently at inference, so that one modality
can receive more loops while the other keeps its depth.

We introduce \textbf{Mixture of Looped Modalities (MoLo)}, which
unties depth across modalities as MoT unties parameters, so that the number of
loops given to each modality, which we call the \emph{allocation}, is set at
inference. The observation behind MoLo is that \textbf{depth, like parameters,
need not be shared across modalities.} Once each modality loops on its own
depth coordinate, one checkpoint can run any allocation, and each modality can
receive the depth it needs. Each modality loops through its own core weights,
and joint attention lets an advancing modality read the latest state of the
others. \Cref{sec:background,sec:molo} describe MoLo, and \cref{sec:experiments} evaluates it by measuring how the loops of each modality affect quality, how to spend an inference budget and how MoLo compares with published generators. \Cref{sec:related_work,sec:conclusion} discuss related work and conclude.

\begin{bluebox}{Contributions}
\begin{list}{\arabic{enumi}.}{%
  \usecounter{enumi}%
  \setlength{\leftmargin}{1.4em}\setlength{\labelwidth}{.9em}\setlength{\labelsep}{.5em}%
  \setlength{\topsep}{0pt}\setlength{\partopsep}{0pt}\setlength{\itemsep}{1pt}\setlength{\parsep}{0pt}}
\item \textbf{One loop per modality.} We introduce MoLo, which gives each
modality its own recurrent core and its own number of loops, so modalities
can run at different depths inside one network. Because training draws each
modality's depths independently, a single checkpoint runs any allocation at
inference, without retraining (\cref{sec:molo}).

\item \textbf{Depth pays off in the generated modality.} More loops improve
whatever is being generated, while the condition needs at most one. Image
loops take CC12M FID from \textbf{31.55 to 13.56}, past the four used in
training, and one text loop more than doubles text rendering over tied
depth. Captions need no image loops at all, which cuts their cost to about
\textbf{a third} (\cref{sec:q1-depth}).

\item \textbf{A compact model that outperforms much larger generators.} A single checkpoint with
188M trainable parameters generates both images and captions. On five
prompt-following benchmarks it averages 66.1, above \textbf{13 of 15}
published text-to-image generators, including all four above 6B parameters,
while using \textbf{under 3\% of their parameters}
(\cref{sec:q4-comparison}).
\end{list}
\end{bluebox}

\section{Background}
\label{sec:background}

MoLo combines three existing components, each meeting one requirement. \textbf{(1)}~All modalities, here image and text, should share one generative objective, as flow matching with endpoint prediction allows. \textbf{(2)}~The passes of a shared core must stay useful at depths not seen in training, which \citet{derakhshani2026lift} report for their Loop Flow Transformer (LiFT); MoLo also uses its depth coordinate to order the passes of different modalities. \textbf{(3)}~Each modality needs its own weights while interacting with the others, as in Mixture-of-Transformers~\citep{liang2024mot}.

\paragraph{(1) Flow matching with endpoint prediction.}
\label{sec:bg-flow-matching}

Flow matching learns a time-dependent velocity field that transports samples from a tractable base distribution $\rho_0$ to the data distribution $\rho_1$~\citep{lipman2023}. Given independent draws $x_0\sim\rho_0$ and $x_1\sim\rho_1$, rectified flow and stochastic interpolants~\citep{liu2023rectified,albergo2023interpolants} connect them by the linear path $x_t=(1-t)x_0+t x_1$ with constant velocity $u^\star=x_1-x_0$, so that $t=0$ corresponds to pure noise and $t=1$ to clean data. Regressing $u^\star$ from $(x_t,t)$ recovers the marginal velocity, and sampling integrates the learned field from $t=0$ to $t=1$ as a neural ODE~\citep{chen2018neural}.

A network can instead predict the clean endpoint $\hat x_1$: since $x_1-x_t=(1-t)(x_1-x_0)$, this implies the velocity $(\hat x_1-x_t)/(1-t)$, and the squared velocity error equals the squared endpoint error weighted by $(1-t)^{-2}$. Because this weight diverges at the clean endpoint, we cap it and train with
\begin{equation}
  \mathcal L_{\mathrm{FM}}
  =\mathbb E_{x_0,x_1,t}\!\left[
    \frac{1}{d\,\max(1-t,\epsilon)^2}\,\|\hat x_1(x_t,t)-x_1\|_2^2\right],
  \label{eq:bg-endpoint-loss}
\end{equation}
where $d$ is the number of scalar coordinates and $\epsilon>0$ sets the cap. Predicting the endpoint also lets the flow run directly in pixel space, without a VAE. Clean images lie near a low-dimensional manifold whereas noise and velocity do not, so a transformer that predicts the clean image can denoise large pixel patches on which noise and velocity prediction fail~\citep{li2026jit}. The same objective extends to text once text is represented continuously: Embedded Language Flows (ELF)~\citep{elf2026} runs flow matching on contextual embeddings from a pretrained text encoder, with the same endpoint prediction and velocity-space loss.

For conditional generation the network also receives a condition $c$, and classifier-free guidance~\citep{ho2022classifierfree} combines its prediction with an unconditional one, obtained by replacing $c$ with a learned null condition,
\begin{equation}
  \tilde x_1=\hat x_1(x_t,t,\varnothing)+g\,\big(\hat x_1(x_t,t,c)-\hat x_1(x_t,t,\varnothing)\big),
  \label{eq:bg-guidance}
\end{equation}
and integrates the velocity of the guided endpoint $\tilde x_1$. The guidance scale $g$ trades diversity for fidelity: $g=1$ disables guidance and needs one network evaluation per step, whereas $g>1$ needs two.

\paragraph{(2) Loop Flow Transformers.}
\label{sec:bg-lift}

A looped transformer applies one shared group of blocks, its core, several times, so that executed depth grows without adding parameters~\citep{dehghani2019universal,geiping2025recurrent}. The Loop Flow Transformer (LiFT) of \citet{derakhshani2026lift} trains such a core for flow-based generation by giving each pass its own target. Within one network evaluation, a prelude produces an initial state $h_0$ whose readout gives an initial prediction $u_0$, which LiFT detaches as an anchor, $b=\operatorname{sg}(u_0)$, and connects to the flow-matching target through a path indexed by a continuous depth coordinate $s$,
\begin{equation}
  \bar u_s=(1-s)\,b+s\,u^\star,\qquad s\in[0,1].
  \label{eq:bg-lift-path}
\end{equation}

The depth coordinate tells each pass how far along this path it should be. During training, LiFT samples intermediate coordinates uniformly, conditions each pass of the core on its coordinate and supervises the readout after that pass at the corresponding point of \cref{eq:bg-lift-path}. The final pass always reaches $s=1$ and is therefore trained on the ordinary flow-matching objective. Because $s$ is continuous, the loop count at inference only sets how finely the path is traversed, and \citet{derakhshani2026lift} report that a LiFT checkpoint keeps improving when run with more loops than it was trained with. MoLo applies the same path to endpoint rather than velocity predictions. 
LiFT, however, generates a single modality and therefore needs only one depth coordinate.

\paragraph{(3) Mixture-of-Transformers.}
\label{sec:bg-multimodal}

With more than one modality, a transformer must also decide which weights process each token. In MMDiT blocks~\citep{esser2024sd3}, image and text tokens have their own projections, feed-forward layers and normalization, and Mixture-of-Transformers (MoT)~\citep{liang2024mot} extends this to every non-embedding parameter of a multimodal transformer. Attention, however, remains joint. Each modality $m\in\mathcal M$ projects its states $H_m$ to its own queries, keys and values and attends over those of all modalities,
\begin{equation}
  O_m=\operatorname{softmax}\!\Big(\frac{Q_mK^\top}{\sqrt{d_{\mathrm h}}}\Big)V,\qquad
  K=\big[H_{m'}W^{K}_{m'}\big]_{m'},\qquad
  V=\big[H_{m'}W^{V}_{m'}\big]_{m'},
  \label{eq:bg-joint-attention}
\end{equation}
where $Q_m=H_mW^{Q}_m$ are the queries of modality $m$, $[\cdot]_{m'}$ concatenates over all modalities along the sequence and $d_{\mathrm h}$ is the head width; modality-specific output projections and feed-forward layers follow. Yet every MoT block updates every modality, so the modalities share one depth. MoLo keeps these blocks but decides at each pass which modalities are updated at what depth. %

\section{Mixture of Looped Modalities}
\label{sec:molo}

\begin{figure}[!htb]
\centering
\includegraphics[width=.95\linewidth]{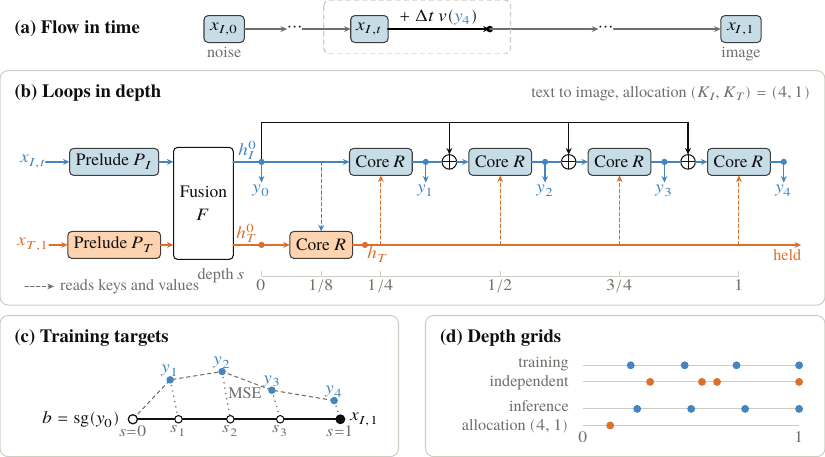}
\caption{\textbf{Each modality loops on its own depth grid,} shown here for text-to-image generation.
(a)~Each integration step evaluates the network in (b).
(b)~Image and text loop through one core $R$ with their own weights, here four times for the image and once for text. Each pass reads the latest state of the other modality (dashed arrows).
(c)~Each estimate $y_k$ is trained toward its point on the path from $\operatorname{sg}(y_0)$ to the clean image.
(d)~Independently sampled grids let one checkpoint run any allocation.}
\label{fig:molo-overview}
\end{figure}

Mixture of Looped Modalities (MoLo, \cref{fig:molo-overview}) turns a stack of Mixture-of-Transformers blocks into a recurrent core and gives each modality its own loop and corresponding depth. As in MoT, each modality has its own weights within the core. Unlike MoT, each modality also has its own depth grid, which sets how many passes it takes through its core. A modality can take several passes while the others wait. At each pass, it attends to the latest state of all modalities through joint attention (\cref{eq:bg-joint-attention}). We define the streams and their initial states (\cref{sec:molo-streams}), order their passes by depth (\cref{sec:molo-recurrence}), train on independently sampled grids (\cref{sec:molo-objective}) and choose the allocation at inference (\cref{sec:molo-inference}).

\subsection{Modality Streams and Their Initial States}
\label{sec:molo-streams}

MoLo treats each modality as a stream that either conditions the generation or is generated. Let $\mathcal M$ index the streams of a task, with one stream $g\in\mathcal M$ to be generated, and let $x_{m,1}$ be the clean continuous representation of modality $m$ in a paired sample. Because the same representation serves as an observation when $m$ conditions another modality stream and as the endpoint when $m$ is generated, one network can learn several conditional tasks by choosing which stream to noise~\citep{bao2023unidiffuser}. During training at generative time $t_g$, the generated stream receives $x_{g,t_g}=(1-t_g)x_{g,0}+t_g x_{g,1}$, with base noise $x_{g,0}$, whereas each conditioning stream enters clean, at $t_m=1$. Generative times stay fixed within one network evaluation (\cref{fig:molo-overview}a). Within the evaluation, each modality stream advances along its own depth coordinate $s_m\in[0,1]$, the counterpart of LiFT's coordinate (\cref{eq:bg-lift-path}).

Each evaluation begins with a private prelude $P_m$ per modality stream, after which joint fusion blocks $F$~\citep{esser2024sd3} update all streams together, so that they exchange information before any loop begins (left of \cref{fig:molo-overview}b). Because the readout head $D_g$ of the generated stream follows fusion, $y_0$ already depends on the conditioning streams when it anchors supervision for the later passes.

\begin{bluebox}{Initial states}
With $t=(t_m)_{m\in\mathcal M}$ and task $\tau$, one network evaluation starts from
\begin{equation}
  (h_m^0)_{m\in\mathcal M}=F\big((P_m(x_{m,t_m};t,\tau))_{m\in\mathcal M}\big),
  \qquad y_0=D_g(h_g^0),
  \label{eq:molo-initial}
\end{equation}
\end{bluebox}

\subsection{Recurrence on Independent Depth Coordinates}
\label{sec:molo-recurrence}

The recurrent core $R$ is a stack of Mixture-of-Transformers blocks. Each modality stream $m$ receives an ordered grid $\mathcal S_m=(s_{m,1},\ldots,s_{m,K_m})$ in $(0,1]$, whose length $K_m$ is the number of loops that stream takes. Together, the loop counts $(K_m)_{m\in\mathcal M}$ form the \emph{allocation}. The generated stream's grid always ends at $s=1$, where its prediction targets the clean endpoint, whereas a conditioning stream's grid may be empty. Since all grids share one interval, they also order the passes of different streams.

\begin{bluebox}{Depth-ordered events}
Let $\sigma_1<\cdots<\sigma_J$ be the distinct depths in the union of the grids. Event $j$ applies one pass of $R$ to the active modality streams $\mathcal A_j=\{m\in\mathcal M:\sigma_j\in\mathcal S_m\}$,
\begin{equation}
  (h_m)_{m\in\mathcal A_j}\leftarrow R\big((\operatorname{Norm}(h_m)+h_m^0)_{m\in\mathcal A_j}\,\big|\,(h_m)_{m\notin\mathcal A_j}\big),
  \label{eq:molo-events}
\end{equation}
in which only the active streams form queries, while every stream supplies keys and values.
\end{bluebox}

The reinjection of $h_m^0$ follows LiFT~\citep{derakhshani2026lift}, and each active modality stream conditions the blocks of $R$ on its new depth $\sigma_j$, the generative times and the task. A waiting stream projects its keys and values from its latest state, conditioned on the depth that state reached, or from $h_m^0$ if it has not yet moved.

Streams at a common depth update together from the same incoming states, whereas a later event reads the result of an earlier one, so the order of events determines what each pass can see. Because $R$ reuses its parameters across events, the grids span a family of computations that one checkpoint can execute. Equal grids give \emph{tied depth}, a looped Mixture-of-Transformers in which every stream updates in every pass. At the other extreme, an empty conditioning grid leaves the condition at $h_m^0$, and in between, a single conditioning pass before the first generated pass lets every generated pass read one refined condition (\cref{alg:molo-evaluation} gives the full procedure).

\subsection{Supervision Across Independent Depths}
\label{sec:molo-objective}

For one checkpoint to serve every allocation, training must expose $R$ to many orderings of events. For each training batch, every stream $m$ therefore draws $K_m-1$ depths uniformly from $(0,1)$, sorts them and appends $s=1$. Because the draws are independent, the passes interleave differently from batch to batch, so a pass may read another stream that is ahead of it or still at depth zero.

Only the generated stream receives direct supervision. Following LiFT, its estimate $y_k$ after the $k$th pass is trained toward a point on the path from the detached initial estimate $b=\operatorname{sg}(y_0)$ to the clean endpoint. All passes share this anchor, including those that read newly updated conditioning states, so a change in the condition alters the prediction but not the path against which it is measured.

\begin{bluebox}{Trajectory supervision}
With the target $\bar y(s)=(1-s)b+s\,x_{g,1}$ at depth $s$, the weight $w(t_g)=\max(1-t_g,\epsilon)^{-2}$ of \cref{eq:bg-endpoint-loss} and $d_g$ target coordinates, the per-example loss is
\begin{equation}
  \mathcal L=\frac{w(t_g)}{d_g}\Big(\frac{1}{K_g}\sum_{k=1}^{K_g}\big\|y_k-\bar y(s_{g,k})\big\|_2^2
  +\lambda\,\big\|y_0-x_{g,1}\big\|_2^2\Big).
  \label{eq:molo-trajectory}
\end{equation}
\end{bluebox}

Since the final pass sits at $s_{g,K_g}=1$, its term regresses the clean endpoint under the same time weighting as ordinary endpoint flow matching. The $\lambda$ term trains the initial estimate and, as in LiFT, keeps the anchor from drifting. The conditioning streams have no target of their own, so within a task they are trained only to be useful to the generated stream, through the keys and values it reads.

\subsection{Inference with a Chosen Allocation}
\label{sec:molo-inference}

At inference, the grids are chosen rather than sampled, and the allocation becomes a deployment parameter. Each stream spaces its passes uniformly at $s_{m,k}=k/K_m$, so consecutive passes of the generated stream aim at equally spaced points on its path and its last pass reaches the clean endpoint. A single conditioning pass is the exception. It goes at $s=1/(2K_g)$, halfway to the first generated pass, so the condition is refined once after fusion and every generated pass reads the refined state. One network evaluation thus runs the preludes and fusion once and then the events of the allocation in depth order (\cref{eq:molo-events}). At each Euler step, the final estimate $y_{K_g}$ gives the velocity $(y_{K_g}-x_{g,t_g})/(1-t_g)$ that advances the generated stream (\cref{sec:bg-flow-matching}), and the next step repeats the evaluation with the same allocation. With guidance, the evaluation runs a second time under the null condition, and the two endpoints are combined as in \cref{eq:bg-guidance}.

\section{Experiments}
\label{sec:experiments}

Because the number of loops per stream is chosen at inference, every experiment evaluates the same checkpoint under different allocations. We first test whether the generated stream keeps improving with more loops, including beyond the four used in training (\textbf{Q1}), and how many loops the conditioning stream needs (\textbf{Q2}). We then fix the conditioning stream at that number and ask how a fixed compute budget is best split between loops, guidance and integration steps, for both text-to-image generation and captioning (\textbf{Q3}). Finally, we compare the resulting model, with 188M trainable parameters, against published generators of up to 6.9B parameters (\textbf{Q4}).

Four results stand out. As in LiFT, extra loops at inference improve the generated stream beyond the training depth, images more than captions. The conditioning stream needs at most one loop, since one text loop gives the best images and captions need no image loops. Guidance and loops are complementary, since the best CMMD and the best captions both combine them, although with strong guidance extra image loops no longer lower FID. At one fixed allocation, the 188M-parameter model has a higher prompt-following average than 13 of 15 published text-to-image generators.

\subsection{Experimental Setup}
\label{sec:exp-setup}

\paragraph{Model.}
\label{sec:model}
We instantiate MoLo with an image stream and a text stream, $\mathcal M=\{I,T\}$, and train one generator for both text-to-image synthesis ($g=I$) and image captioning ($g=T$), writing allocations as $(K_I,K_T)$, image loops first. The image stream works directly on $512\times512$ RGB pixels. The text stream works on the hidden states of a frozen FLAN-T5-Large encoder~\citep{chung2024flant5}, which a frozen FLAN-T5 decoder turns back into tokens after captioning. Each stream has four prelude blocks, two fusion blocks and a six-block core, so the generator stores 12 blocks per stream yet executes 30 per network evaluation at its four training loops. It has 188M trainable parameters, excluding the 341M-parameter frozen encoder (\cref{sec:app-architecture}).

\paragraph{Training.}
We pretrain for one million updates on CC12M~\citep{changpinyo2021cc12m} and FLUX-Reason~\citep{fang2026prism} and fine-tune for 160,000 updates on a mix of curated image--caption sets (Fine-T2I, BLIP3-o-60K, DALL-E~3 Discord and ShareGPT-4o-Image; \cref{sec:app-training}). Both stages sample the two directions with equal probability, train each stream with four loops and drop the condition in 10\% of examples, which enables classifier-free guidance~\citep{ho2022classifierfree} in both directions.

\paragraph{Inference.}
All experiments use the EMA weights of the final checkpoint and fix prompts and initial noise across allocations, with the generated stream taking 1 to 16 loops and the conditioning stream 0 to 16 (\cref{sec:app-evaluation}).
Image quality is measured on a fixed bank of 10,000 CC12M prompts and reference images (\cref{sec:app-cc12m-scaling}). Captions are generated for 5,000 COCO~\citep{lin2014coco} photographs at 32 steps and guidance scale 2. Cost is measured as generator FLOPs and as end-to-end latency on one H100.

\begin{figure}[!b]
\centering
\includegraphics[width=\linewidth]{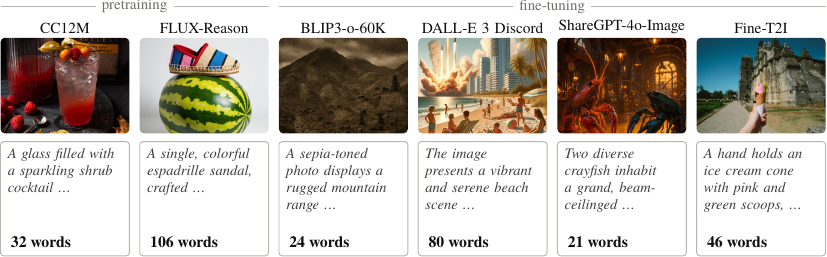}
\caption{\textbf{Training captions are two to ten times longer than COCO's references, so MoLo writes captions that n-gram metrics and CLIPScore cannot score.} One pair per training source, with the opening words and full length of its caption. COCO's references have a median of 10 words.}
\label{fig:training-source-examples}
\end{figure}

\paragraph{Metrics.}
For images we report FID~\citep{heusel2017fid} and CMMD~\citep{jayasumana2024cmmd}, a maximum mean discrepancy with a Gaussian kernel on CLIP image embeddings. We report both because \citet[Sec.~6.1]{jayasumana2024cmmd} find that CMMD keeps decreasing across successive refinements of Muse and Stable Diffusion even where FID suggests that quality worsens, which is the regime that additional loops create. Image--prompt alignment is measured with CLIPScore~\citep{hessel2021clipscore}. MoLo, trained on long descriptions (\cref{fig:training-source-examples}), often writes captions beyond the 77 tokens that CLIPScore reads, so we score captions with Long-CLIP~\citep{zhang2024longclip}. For prompt following we use GenEval~\citep{ghosh2023geneval}, DPG-Bench~\citep{hu2024ella}, PRISM~\citep{fang2026prism}, T2I-CoReBench~\citep{li2026corebench} and SpatialGenEval~\citep{wang2026spatial}.

\subsection{Q1 and Q2: Where Does Loop Depth Help?}
\label{sec:q1-depth}

\begin{question}{Q1}
Does the generated stream keep improving as it receives more loops, even beyond the four used in training?\\[2pt]
\cardlabel{cardink}{Q2}\enspace Once the generated stream has its loops, how many should the conditioning stream receive?
\end{question}

\begin{figure}[!htb]
\centering
\includegraphics[width=\linewidth]{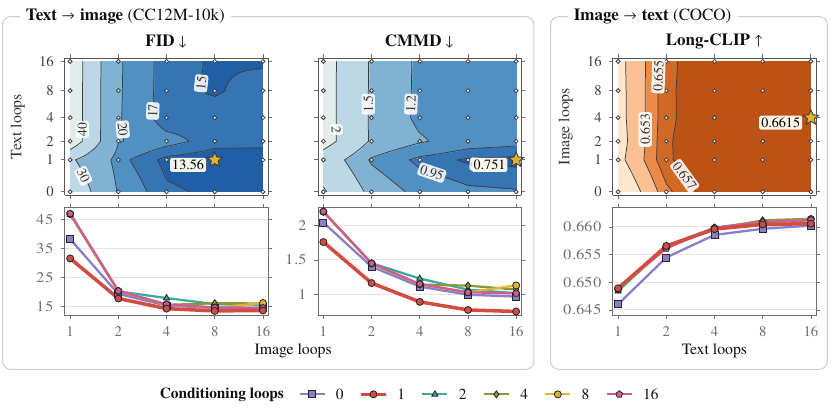}
\caption{\textbf{In both directions, loop depth helps the generated stream, and the conditioning stream needs at most one loop.} Contour maps show each score over the loops of both streams, with the generated stream's loops on the horizontal axis, and the curves below give one series per number of conditioning loops. FID and CMMD use CC12M-10k at 100 integration steps without guidance, and Long-CLIP uses 5,000 COCO photographs at 32 steps and guidance scale 2, with 95\% intervals of about $\pm0.0016$ (\cref{sec:app-evaluation}). Stars mark the best scores.}
\label{fig:cc12m-text}\label{fig:caption-longclip}
\end{figure}

The text-to-image panel of \cref{fig:cc12m-text} answers the question for images along its one-text-loop row. From one to four image loops, FID falls from 31.55 to 14.34 and CMMD from 1.760 to 0.891, and both keep decreasing at eight loops, well beyond the training depth. At sixteen loops CMMD falls further, from 0.774 to 0.751, while FID changes by only 0.15 (\cref{sec:app-cc12m-scaling}).

\Cref{fig:image-progression}(a) shows what these gains look like when only the image loops change. A thank-you sign is spelled correctly from eight loops, a face moves into the requested small screen, and two traffic lights become the requested four, turning GenEval's verdict from fail to pass. Lettering improves most visibly, as in \cref{fig:cover,fig:loop-visual}.

Captions improve along the horizontal axis of the image-to-text panel, mostly by the training depth. With the photograph read only through fusion, Long-CLIP rises from 0.6460 at one text loop to 0.6585 at four and changes little beyond. Individual captions still change past four loops (\cref{fig:caption-main}(b)): the kitchen's appliances become an oven and a stove from eight loops, and the skier's caption gains the visible ski pole at sixteen. At eight integration steps and guidance scale 6, the most efficient setting in \cref{sec:q3-envelope}, Long-CLIP changes little beyond two text loops (\cref{tab:caption-default-grid}).

\definecolor{panelrule}{HTML}{CECDC3}
\newcommand{\samplepanel}[2]{%
  \begin{tikzpicture}
  \node[minimum width=\linewidth-1pt,inner xsep=0pt,inner ysep=6pt] (p) {\includegraphics[width=.88\linewidth]{#2}};
  \draw[panelrule,line width=.5pt,rounded corners=4pt] ([yshift=6pt]p.north west) rectangle (p.south east);
  \node[fill=white,inner xsep=3pt,anchor=west,font=\rmfamily\bfseries\footnotesize]
    at ([xshift=8pt,yshift=6pt]p.north west) {#1};
  \end{tikzpicture}}
\begin{figure}[!tp]
\centering
\samplepanel{(a) Text $\rightarrow$ image}{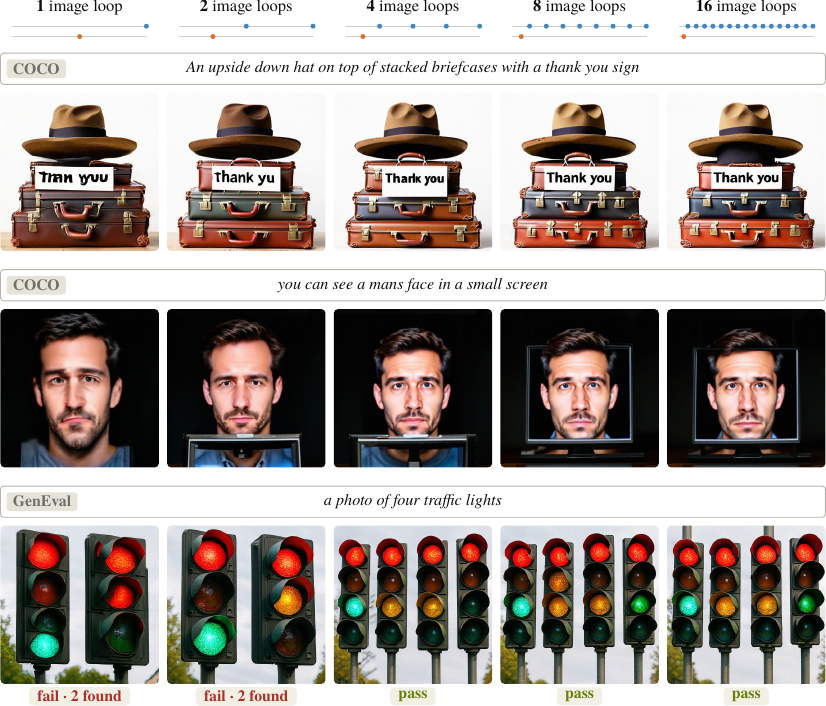}\par\smallskip
\samplepanel{(b) Image $\rightarrow$ text}{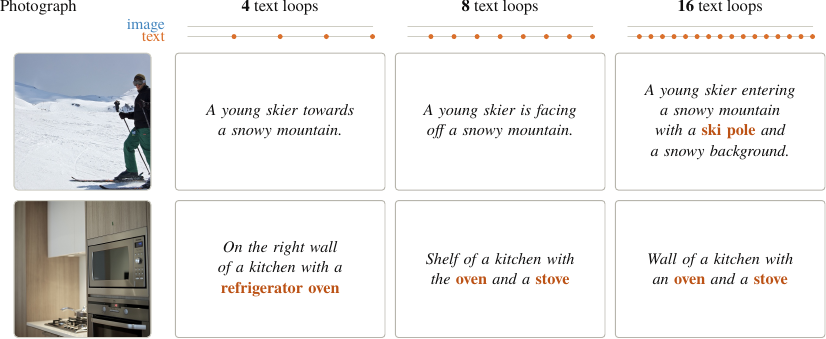}
\caption{\textbf{With prompt and noise fixed, more loops in the generated stream correct what it gets wrong.} (a)~More image loops correct visual spelling, placement and counts. The thank-you sign is spelled correctly from eight loops. The face sits inside the requested small screen from eight loops, and the four traffic lights appear from four loops, where GenEval's verdict turns from fail to pass. (b)~More text loops refine what a caption names, with the photograph read through fusion only. The kitchen's built-in appliances are confused at four text loops and named as an oven and a stove from eight, and the skier's caption adds the visible ski pole at sixteen. The skier row uses 32 integration steps and guidance scale 2, and the kitchen row eight steps and guidance scale 6. Rails above each column mark where the image (blue) and text (orange) passes fall in depth.}
\label{fig:image-progression}\label{fig:caption-main}
\end{figure}

The vertical axes of \cref{fig:cc12m-text} answer the other half of the question. For images, the best scores at every image depth lie on the row with one text loop. Removing that loop brings the scores back close to tied depth, even though the prompt still reaches the image through fusion, so the gain comes from refining the condition once before the image passes read it. This pass reads the image's state after fusion, so it refines the prompt against the image being generated, unlike a prompt encoded once before generation. Further text loops hurt, and most when the image is shallow.\footnote{With more than one loop, the text passes are spread over depth like the image passes (\cref{sec:molo-inference}), so a schedule that places several text passes before the first image pass remains untested.} At one image loop, a second text loop raises FID from 31.55 to 46.62. With three fewer text loops, $(4,1)$ also improves on the training allocation $(4,4)$ at lower compute, and the prompt-following benchmarks of \cref{sec:q4-comparison} agree: one text loop scores above tied depth on every benchmark at four, eight and sixteen image loops (\cref{sec:benchmark-allocation-details}). The gain is largest in text rendering, where T2I-CoReBench's score more than doubles over tied depth at every image depth, from 13.3 to 32.0 at four image loops (\cref{tab:corebench-categories}).

For captions, the vertical axis is nearly flat, and at sixteen text loops all six image depths lie within overlapping 95\% intervals. The photograph still shapes the caption through fusion, as a shuffle control confirms (\cref{sec:caption-results}). Because image loops are the expensive ones (\cref{sec:app-architecture}), this flat direction is also the costly one: dropping all image loops cuts the cost of a caption to about a third, from 151 to 52 TFLOPs, with no measurable change in Long-CLIP.

The right depth for the condition therefore depends on the direction. A prompt conditions images best with one loop, whereas a photograph conditions its caption best with none. The generated streams differ as well, with images improving through eight loops and captions leveling off near four, and the best image allocation changes from benchmark to benchmark (\cref{tab:external-comparison}, last row). No tied depth, and no fixed rule such as conditioning once, matches all of these, whereas a per-modality allocation chosen at inference can.

\begin{finding}
More loops improve the generated stream, images more than captions. Without guidance, image distribution scores improve through eight image loops, beyond the training depth, whereas caption alignment levels off after about four text loops. The conditioning stream needs at most one loop, one as text and none as an image. One text loop gives the best image scores at every image depth, whereas captions need no image loops.
\end{finding}

\subsection{Q3: How Should an Inference Budget Be Spent?}
\label{sec:q3-budget}

\begin{question}{Q3}
At a fixed compute budget, how should the loops of the generated stream be traded against guidance and integration steps?
\end{question}

With the conditioning stream settled, each direction has three remaining controls at inference: the depth of the generated stream, the guidance scale and the number of integration steps. All three cost compute, and they interact. We examine guidance and integration steps in turn, and then all three under a fixed budget.

\subsubsection{Guidance}
\label{sec:q3-guidance}

\begin{figure}[!tbp]
\centering
\includegraphics[width=\linewidth]{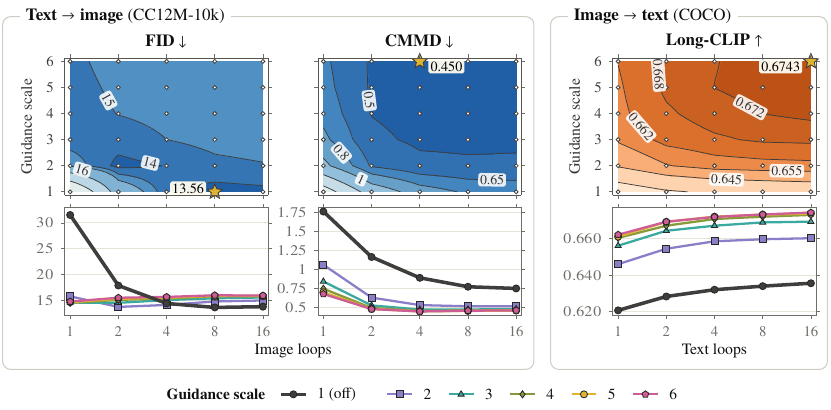}
\caption{\textbf{Scores over the loops of the generated stream and the guidance scale.} The panels are laid out as in \cref{fig:cc12m-text}. For images, the best CMMD combines guidance with four image loops (0.450), although with guidance FID changes little with image depth. For captions at 32 steps, the two add up, with sixteen text loops scoring highest at every scale and scale 6 giving the best Long-CLIP (0.6743). FID and CMMD use CC12M-10k at one text loop and 100 integration steps, where scale 1 disables guidance. Long-CLIP uses 5,000 COCO photographs, with zero image loops and 32 integration steps. Stars mark the best scores.}
\label{fig:cc12m-cfg}
\end{figure}

FID and CMMD disagree about guidance (\cref{fig:cc12m-cfg}). With strong guidance, FID is lowest at a single image loop, rising at scale 6 from 14.71 at one loop to 15.93 at eight. Without guidance, it is lowest at eight image loops (13.56). CMMD favors guidance at every image depth and is lowest at four loops with scale 5 or 6 (0.450). This fits the usual reading that guidance trades diversity, which FID rewards, for fidelity~\citep{ho2022classifierfree}. At eight image loops, scale 6 also raises CLIPScore from 0.701 to 0.765. At high image depth, disabling guidance therefore gives the best FID and saves the unconditional network evaluation at every integration step.

\begin{figure}[!tbp]
\centering
\includegraphics[width=\linewidth]{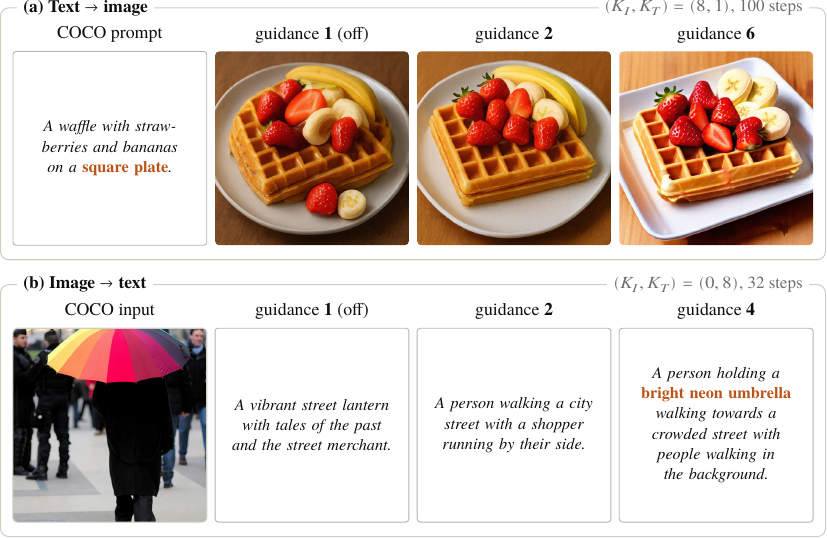}
\caption{\textbf{Guidance changes which details the model captures at fixed depth.} (a)~At $(K_I,K_T)=(8,1)$ and 100 integration steps, the requested square plate appears at guidance scale 6. (b)~At $(0,8)$ and 32 steps, the caption describes a street lantern with guidance disabled and identifies the umbrella at scale 4. Each row uses the same initial noise across guidance scales.}
\label{fig:sampling-cfg}
\end{figure}

\Cref{fig:cc12m-cfg} (right) and \cref{tab:caption-budget-guidance} show Long-CLIP for captions. At 32 steps, guidance at scale 4 raises a single text loop from 0.6208 to 0.6604. Sixteen text loops without guidance reach only 0.6356. At this step count the two effects are close to additive, with sixteen text loops scoring 0.0121 to 0.0148 above one loop at every scale. At eight steps they are not, as sixteen text loops add 0.0071 without guidance but only 0.0018 at scale 6 (\cref{tab:caption-budget-guidance-steps}). \Cref{fig:sampling-cfg} shows how guidance changes a single image and caption at fixed depth.

\begin{finding}
At high image depth, omitting guidance minimizes FID and saves an unconditional pass, whereas CMMD and CLIPScore still favor it. For captions, guidance and text loops add up. At 32 steps, sixteen text loops raise Long-CLIP at every guidance scale, and the best caption setting combines strong guidance with four text loops at eight steps.
\end{finding}

\subsubsection{Integration Steps}
\label{sec:q3-steps}

\begin{figure}[!tbp]
\centering
\includegraphics[width=\linewidth]{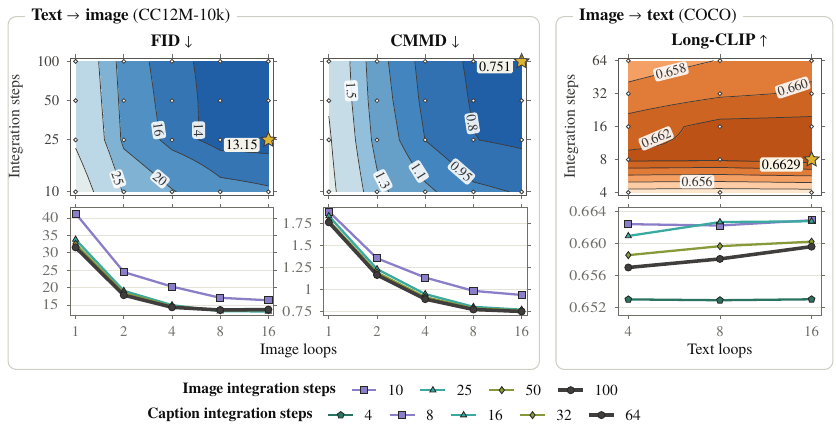}
\caption{\textbf{Up to 25 integration steps, more steps and more image loops both help images, whereas captions peak at eight to 16 steps.} Scores over the loops of the generated stream and the number of integration steps, laid out as in \cref{fig:cc12m-text}. FID and CMMD use CC12M-10k at one text loop without guidance. Stars mark the best scores. Long-CLIP uses 5,000 COCO photographs with zero image loops and guidance scale 2. The caption panel shows the three text depths evaluated at all five step counts.}
\label{fig:cc12m-steps}
\end{figure}

\begin{figure}[!tbp]
\centering
\includegraphics[width=\linewidth]{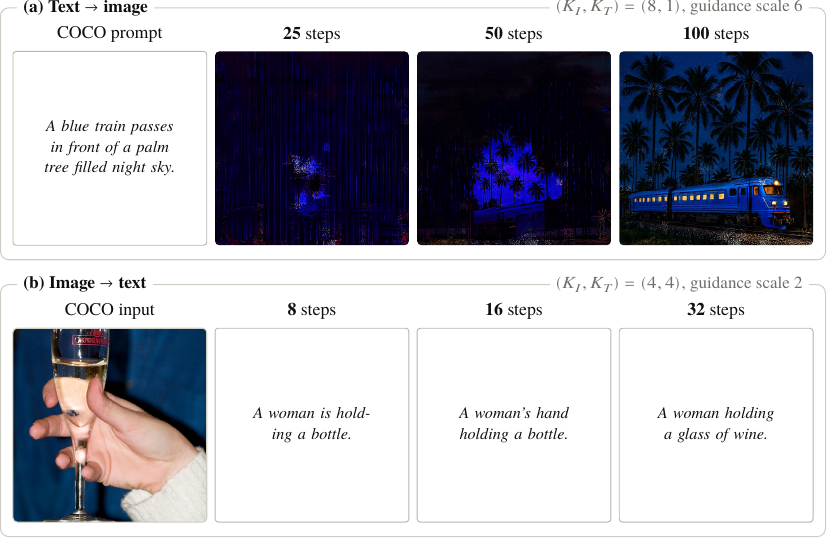}
\caption{\textbf{At fixed depth, more integration steps still change the sample.} (a)~The train and the palms become clearly visible only at 100 steps. (b)~The caption calls the wine glass a bottle at 8 and 16 steps and names it at 32. Each row keeps its initial noise.}
\label{fig:sampling-main}
\end{figure}

Integration steps and loops refine different things: a step advances the image along its flow, whereas the loops refine the endpoint prediction that the step uses. In the text-to-image panel of \cref{fig:cc12m-steps}, both help at every image depth up to 25 steps, but a loop buys far more than a step. One image loop at 100 steps reaches an FID of 31.55 and a CMMD of 1.760 at 29.4 TFLOPs per image, whereas four loops at 25 steps reach 14.97 and 0.948 at 16.0 TFLOPs. Beyond 25 steps the metrics part ways at high image depth: at eight and sixteen image loops, FID is lowest at 25 steps, while CMMD keeps decreasing through 100. At a fixed allocation, extra steps also change what a single sample shows (\cref{fig:sampling-main}).

Captions need far fewer steps than images (\cref{fig:cc12m-steps}, right, and \cref{tab:caption-budget-steps}). Long-CLIP peaks at eight or 16 steps for every text depth tested, at guidance scale 2 as at scale 6. At scale 2, halving from eight to four steps lowers it by 0.0094 to 0.0099, and beyond 16 steps it falls again, to 0.6570 at 64 steps with four text loops. \Cref{fig:caption-step-examples} compares individual captions at the two step counts.

\begin{finding}
A loop buys far more than an integration step: four loops at 25 steps improve on one loop at 100 steps in both metrics with 45\% less compute. At high image depth, FID needs only 25 steps, whereas CMMD keeps improving through 100. Captions need fewer steps than images, with Long-CLIP highest at eight to 16.
\end{finding}

\subsubsection{Budget}
\label{sec:q3-envelope}

\begin{figure}[!htbp]
\centering
\includegraphics[width=\linewidth]{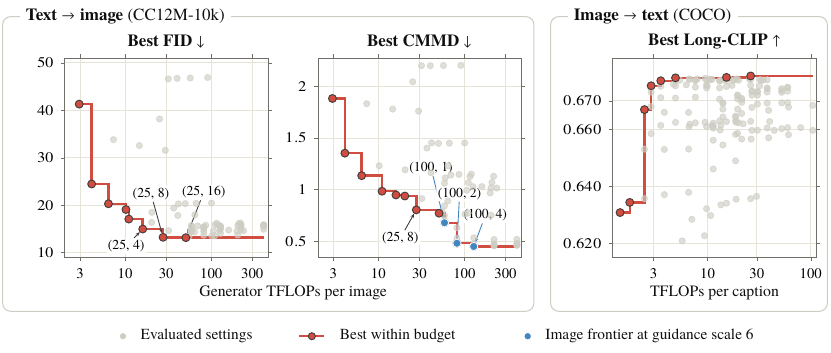}
\caption{\textbf{Best observed score within each generator budget.} The image panels use 70 CC12M-10k settings, and the caption panel uses 153 settings on 5,000 COCO photographs. Gray dots mark evaluated settings. The red staircases track the best score among settings at or below each cost. Blue dots mark image frontier points with guidance scale 6. Labels in the image panels give the integration steps and image loops of each point on the envelopes. \Cref{tab:caption-budget-frontier} gives the settings on the caption frontier.}
\label{fig:cc12m-compute}
\end{figure}

For images, a fixed budget goes furthest when spent on loops (\cref{fig:cc12m-compute}). Up to 50.8 TFLOPs per image, both envelopes follow unguided settings at 10 or 25 steps, and as the budget grows the best of them moves from one to sixteen image loops. Eight loops at 25 steps already reach an FID of 13.23, below the best guided FID at a third of its cost. Guidance enters the envelopes only beyond that budget, and only for CMMD, which guided settings at 100 steps lower from 0.772 to 0.450.

Captions level off at a far smaller budget, below 5 TFLOPs per caption (\cref{fig:cc12m-compute}, right, and \cref{tab:caption-budget-frontier}). There, four text loops at eight steps and guidance scale 6 reach a Long-CLIP of 0.6781, above every 32-step setting, and costlier settings on the envelope add at most 0.0006, inside the 95\% interval. Against the 32-step sampler at guidance scale 2 that \cref{sec:q1-depth} uses to compare allocations, this setting scores 0.0196 higher at the same four text loops and halves the time per caption on an H100, from 3.78 to 1.76 seconds.

\begin{finding}
For images, a fixed budget is best spent on image loops, run at few steps and without guidance. This gives the best FID at every budget and the best CMMD up to moderate budgets, after which guidance pays off. Captions come within 0.0006 of their best Long-CLIP with four text loops at eight steps and guidance scale 6, at 4.9 TFLOPs and without image loops.
\end{finding}

\FloatBarrier
\subsection{Q4: How Does MoLo Compare with Larger Generators?}
\label{sec:q4-comparison}

\begin{question}{Q4}
How does a fixed MoLo allocation compare with published text-to-image generators?
\end{question}

\begin{figure}[!t]
\centering
\includegraphics[width=\linewidth]{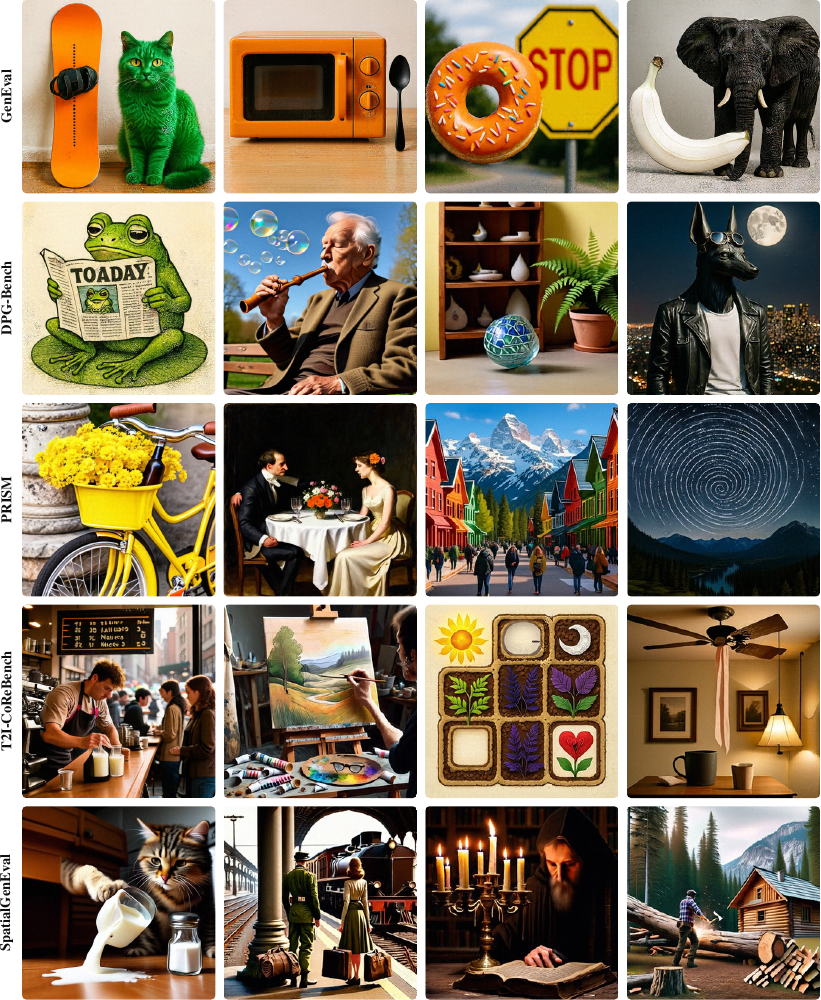}
\caption{\textbf{MoLo at $(8,1)$ on the five benchmarks of \cref{tab:external-comparison}.} One row per benchmark.}
\label{fig:comparison-gallery}
\end{figure}

\Cref{tab:external-comparison} places MoLo among the fifteen published text-to-image generators compared by \citet{chng2026loopeddit}, at one fixed allocation, $(8,1)$, which follows Q1 and Q2 with eight image loops and one text loop, and at the training allocation $(4,4)$. \Cref{fig:comparison-gallery} shows samples at $(8,1)$ for each benchmark.

\begin{table}[!t]
\centering
\newcommand{\molobestcell}[2]{\shortstack{#1\\[-1pt]{\scriptsize #2}}}
\caption{\textbf{With under 3\% of the parameters of the largest models, MoLo has a higher average than 13 of 15 published generators.} Prompt-following scores on five benchmarks, higher is better. Baseline scores are from the comparison of \citet{chng2026loopeddit}, which evaluates each model under its official inference settings. The first two MoLo rows fix the allocation $(K_I,K_T)$ across benchmarks and use 100 integration steps at guidance scale 6. The last row gives the best score MoLo reaches on each benchmark across allocations, with the allocation below. All parameter counts exclude frozen text encoders, such as MoLo's 341M-parameter FLAN-T5 encoder. $^\dagger$Loops one core with tied depth and is trained for text-to-image generation only.}
\label{tab:external-comparison}
\begin{tabularx}{\linewidth}{l*{7}{>{\raggedleft\arraybackslash}X}}
\toprule
& & \multicolumn{5}{c}{Benchmark} & \\
\cmidrule(lr){3-7}
Model & Params & GenEval & DPG & PRISM & CoRe & Spatial & Average \\
\midrule
\multicolumn{8}{l}{\emph{Non-CoT models}}\\
MiniT2I-B/16~\citep{wang2026minit2i} & 0.26B & 87.5 & 84.1 & 55.6 & 44.0 & 52.0 & 64.6 \\
Looped-DiT B/16$^\dagger$~\citep{chng2026loopeddit} & 0.26B & 87.4 & 87.0 & 67.0 & 53.5 & 54.6 & 69.9 \\
E-MMDiT~\citep{shen2025emmdit} & 0.30B & 69.1 & 80.4 & 51.6 & 30.7 & 45.7 & 55.5 \\
DreamLite~\citep{feng2026dreamlite} & 0.39B & 71.6 & 85.2 & 55.5 & 42.2 & 53.9 & 61.7 \\
SANA-0.6B~\citep{xie2025sana} & 0.59B & 65.8 & 83.3 & 56.4 & 39.8 & 47.4 & 58.5 \\
URSA-0.6B~\citep{deng2025ursa} & 0.86B & 64.1 & 85.6 & 59.6 & 43.4 & 52.0 & 60.9 \\
TiM-T2I~\citep{wang2025transition} & 0.87B & 82.8 & 83.2 & 51.4 & 36.7 & 48.2 & 60.5 \\
MiniT2I-L/16~\citep{wang2026minit2i} & 0.91B & 88.3 & 84.8 & 58.9 & 44.4 & 52.2 & 65.7 \\
DeCo-XXL/16~\citep{ma2025deco} & 1.1B & 82.0 & 82.0 & 52.6 & 34.8 & 49.8 & 60.2 \\
UniLiP-3B~\citep{tang2025unilip} & 1.6B & 90.3 & 83.4 & 58.5 & 45.0 & 51.6 & 65.8 \\
InternVL-U~\citep{tian2026internvlu} & 1.7B & 85.0 & 85.2 & 63.5 & 48.2 & 54.5 & 67.3 \\
CogView4~\citep{cogview42025,zheng2024cogview3} & 6.4B & 73.0 & 85.1 & 60.9 & 45.0 & 53.3 & 63.5 \\
\midrule
\multicolumn{8}{l}{\emph{CoT-reasoning models}}\\
Uni-CoT~\citep{qin2025unicot} & 6.6B & 81.2 & 84.1 & 58.8 & 45.8 & 53.8 & 64.7 \\
GoT-R1~\citep{duan2025gotr1} & 6.9B & 72.5 & 84.0 & 56.4 & 42.2 & 50.0 & 61.0 \\
T2I-R1~\citep{jiang2025t2ir1} & 6.9B & 78.8 & 84.7 & 57.1 & 35.5 & 49.7 & 61.2 \\
\midrule
\multicolumn{8}{l}{\emph{Ours}}\\
\rowcolor{findingfill}\textbf{MoLo} $\boldsymbol{(4,4)}$ & \textbf{0.188B} & 84.6 & 83.5 & 56.5 & 45.7 & 50.6 & 64.2 \\
\rowcolor{findingfill}\textbf{MoLo} $\boldsymbol{(8,1)}$ & \textbf{0.188B} & 85.2 & 84.6 & 59.1 & 49.3 & 52.1 & 66.1 \\
\rowcolor{findingfill}\shortstack[l]{\textbf{MoLo}\\[-1pt]{\scriptsize best per benchmark}} & \molobestcell{\textbf{0.188B}}{\phantom{$(2,1)$}}
  & \molobestcell{85.6}{$(2,1)$}
  & \molobestcell{84.8}{$(2,1)$}
  & \molobestcell{59.1}{$(8,1)$}
  & \molobestcell{49.3}{$(8,1)$}
  & \molobestcell{52.7}{$(2,1)$}
  & \molobestcell{66.3}{\phantom{$(2,1)$}} \\
\bottomrule
\end{tabularx}
\end{table}

At $(8,1)$, MoLo averages 66.1 over the five benchmarks, above 13 of the 15 published generators. CogView4 and the three chain-of-thought reasoning models, each with more than 30 times MoLo's parameters, all average lower, and MiniT2I-B/16, the 0.26B model on which Looped-DiT builds, is ahead only on GenEval. Only InternVL-U, with nine times MoLo's parameters, and Looped-DiT average higher. Looped-DiT loops one MMDiT core with tied depth and trains on images alone for 580K updates at the same batch size, about as many as MoLo spends on images while it learns captioning with the same parameters. Because the two models also differ in backbone and supervision, untying is measured within MoLo: the training allocation $(4,4)$, which ties depth, averages 64.2, so $(8,1)$ gains 1.9 points without retraining, and choosing the allocation per benchmark, as in the last row, raises the average to 66.3 (\cref{tab:image-allocation,tab:benchmark-depth-extension}).

\Cref{tab:external-comparison} compares models by parameter count, which sets their memory footprint. At $(8,1)$, 100 steps and guidance scale 6, MoLo spends 221 generator TFLOPs per image. Within MoLo, the cost of every allocation is known, and fewer conditioning loops cost less and score higher. Going from the training allocation $(4,4)$ to $(4,1)$ lowers the cost from 139 to 128 TFLOPs per image and raises the average from 64.2 to 66.0, within 0.1 of $(8,1)$ at 58\% of its compute (\cref{tab:image-allocation}).

\FloatBarrier
\begin{finding}
With 188M trainable parameters and one fixed allocation, MoLo has a higher five-benchmark average than 13 of 15 published generators, including all four above 6B parameters, with under 3\% of their parameters, and the same checkpoint also captions images.
\end{finding}

\clearpage
\subsection{Visualization Across Loops}
\label{sec:visual-loops}

\begin{figure}[!bp]
\centering
\includegraphics[width=\linewidth]{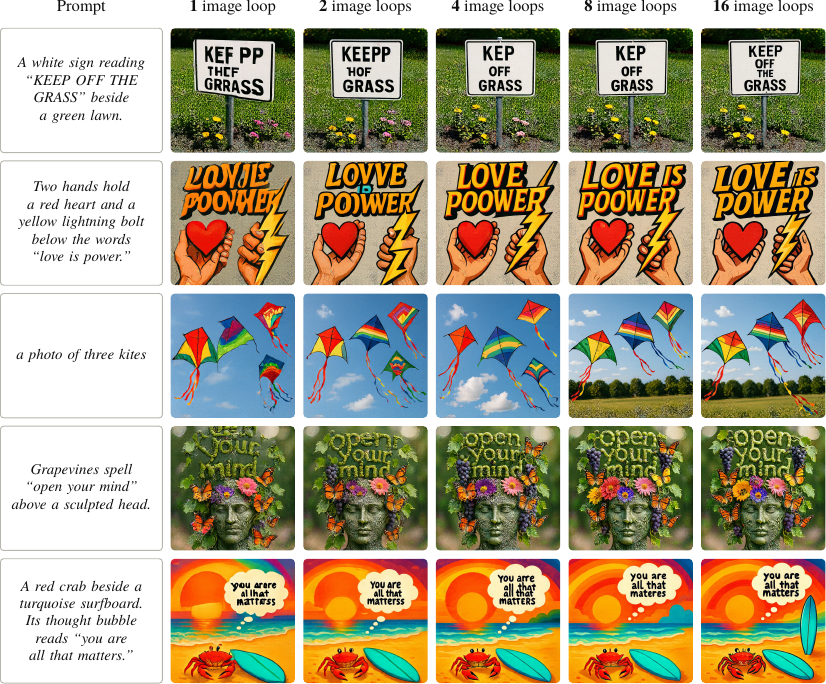}
\caption{\textbf{Image loops change lettering and counts in matched samples.} Five examples from GenEval and DPG-Bench, with the longer prompts abridged. Each row fixes the prompt and initial noise and uses one text loop, 100 integration steps and guidance scale 6. Three kites appear from eight image loops onward. The lawn sign first reads \texttt{KEEP OFF THE GRASS} in full at sixteen loops, and the poster gains its missing \texttt{IS} at eight loops and reads \texttt{LOVE IS POWER} at sixteen. At sixteen loops, the crab's thought bubble spells out the requested phrase.}
\label{fig:loop-visual}
\end{figure}

\Cref{fig:loop-visual} holds the prompt and initial noise fixed in each row. In the poster, the heart and lightning bolt are recognizable at one loop, and the lettering becomes legible as \texttt{LOVE IS POWER} at sixteen. The lawn sign retains its setting and placement while its lettering changes. In the kite example, the extra kite disappears at eight loops. The frequency content of these changes, measured over all GenEval prompts, shows the same shift toward detail (\cref{sec:app-frequency}). Each doubling of image loops changes the image less than the one before, and the share of the change at wavelengths of at most 8 pixels grows from 13\% between one and two loops to 24\% between eight and sixteen.

\begin{figure}[!t]
\centering
\includegraphics[width=\linewidth]{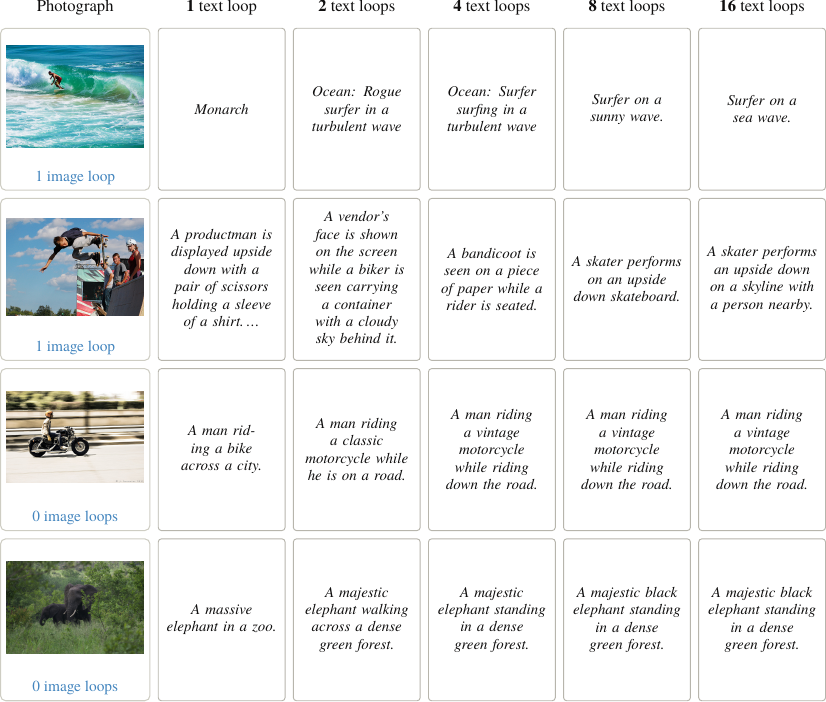}
\caption{\textbf{By sixteen text loops, every caption names its subject.} Each COCO row keeps the photograph, initial noise and number of image loops fixed across columns, with one image loop in the first two rows and none in the last two. Captions use 32 integration steps and guidance scale 2. The surfer and the motorcycle are named from two text loops and the skateboarder from eight, and the elephant's setting changes from a zoo to a green forest at two. The skateboarder's one-loop caption shows its first sentence.}
\label{fig:loop-caption-visual}
\end{figure}

\Cref{fig:loop-caption-visual} shows how the generated caption changes with text depth when the photograph and initial noise are fixed.

\begin{finding}
Additional loops refine the image progressively, most visibly in lettering. Each doubling changes the image less than the one before, and a growing share of that change lies in fine detail.
\end{finding}

\FloatBarrier

\section{Related Work}
\label{sec:related_work}

\paragraph{Modality-specific parameters.}
Joint attention lets modalities exchange information while each keeps its own weights. MMDiT blocks apply separate transformations to image and text tokens and attend over all of them~\citep{esser2024sd3}, and Mixture-of-Transformers extends this to every non-embedding layer, matching dense models with less pretraining compute in several settings~\citep{liang2024mot}. In both, the modality alone selects the weights, without a learned router. Modality-aware mixtures of experts keep this split and add a router inside it. MoMa sends each token to its modality's group of experts, routes it to experts within that group with a learned router, and adds a second router, applied before the split, that decides which tokens each layer processes~\citep{lin2024moma}. In all three, the modality decides which weights process a token, whereas the number of times they are applied is either shared across modalities or routed token by token. MoLo assigns depth by modality as well, the way MoT assigns weights. This includes the depth of the condition, where text-to-image generators sit at two extremes. Cross-attention generators encode the prompt once with a frozen text encoder and update only the image~\citep{saharia2022imagen,chen2024pixartalpha,xie2025sana}, whereas MMDiT updates the prompt in every block. In MoLo the prompt can take any number of loops, and one text loop gives the best images at every image depth (\cref{sec:q1-depth}).

\paragraph{Recurrent depth.}
Reusing blocks makes depth adjustable without changing model size. Universal Transformers bring this recurrence to the transformer~\citep{dehghani2019universal}, and on algorithmic tasks recurrent networks keep computing usefully beyond the iterations seen in training~\citep{schwarzschild2021algorithm,bansal2022extrapolation}. Recurrent language models train over a distribution of recurrence counts~\citep{geiping2025recurrent}. Conditioning the shared block on its depth lets each pass learn a different computation~\citep{xu2025timestep}, and LoopFormer adds the step size to this conditioning so that shorter rollouts reach the representation of a full-depth one~\citep{jeddi2026loopformer}.

Every token in these models receives the same depth. Mixture-of-Depths relaxes this by routing a fixed number of positions through each layer~\citep{raposo2024mod}. $\gamma$-MoD carries the routing to multimodal language models~\citep{luo2025gammamod}, and p-MoD applies it to vision tokens only~\citep{zhang2025pmod}. Mixture-of-Recursions applies the same idea to a shared block, learning which tokens continue through each pass~\citep{bae2025mor}. In each case, a router learned in training decides how deep each token goes, within a capacity or recursion count fixed before training, and in Mixture-of-Depths a token that skips a block also drops out of its attention. RecursiveVLM comes closer to a per-modality design, with modality-specific projections between the recursions of a vision-language decoder~\citep{xu2026recursivevlm}, yet its image and text states still advance on a common schedule.

MoLo allocates depth at a different level and at a different time. Its unit is the modality, since in our experiments the role of a stream sets the depth it needs, with several loops for the generated stream and at most one for the conditioning stream. No router fixes the allocation, so it is chosen at inference. One checkpoint can therefore give a prompt one loop when generating an image, give a photograph none when captioning it, and run the generated stream beyond the four loops seen in training. A stream that waits also stays in joint attention, so the advancing stream reads its latest state at every pass.

\paragraph{Recurrence in generative transformers.}
In a generative transformer, the same blocks can run several times at a fixed noise level before the sampler takes another step. DiT-Air studies layer sharing for text-to-image diffusion~\citep{chen2025ditair}, and LoopDiT combines a prelude, a shared core and an output stage~\citep{chai2026loopdit}. Looped-DiT, the closest architecture to MoLo, loops a shared MMDiT core inside each integration step and regulates the repeated updates with self-modulating attention~\citep{chng2026loopeddit}. A single loop count governs both modalities. Image and text therefore advance together on every pass, and loop depth is traded against integration steps through that one count. Splitting it per modality, as we do, shows that the generated stream needs several loops and the conditioning stream at most one.

\citet{derakhshani2026lift} demonstrate with LiFT that how intermediate passes are supervised decides whether extra loops help at inference. Most methods ask every supervised pass for the final answer. Looped-DiT gives each pass the clean image as its target, and DeepFlow supervises velocity readouts from separate branches of a fixed-depth model~\citep{shin2025deepflow}. Elastic Looped Transformers distill the full-depth prediction into sampled intermediate exits~\citep{goyal2026elt}, and Distill to Loop trains every pass of a recurrent student toward a dense teacher's prediction~\citep{wang2026distilltoloop}. LiFT instead gives each readout a depth-dependent target on the path from the detached initial prediction to the flow-matching target~\citep{derakhshani2026lift}, and MoLo applies this supervision to its generated stream.

Hidden states can also carry computation across integration steps. Recurrent Interface Networks pass latent tokens from one denoising evaluation to the next~\citep{jabri2023rin}, and Thinking with Looped Flows couples a recurrent state to the evolving flow trajectory~\citep{suleymanzade2026thinking}. MoLo rebuilds its states at every integration step instead, so depth can change while the number of steps stays the same.

\paragraph{Continuous image and text generation.}
Unified generators differ first in how they represent text. Transfusion~\citep{zhou2024transfusion} and Show-o~\citep{xie2024showo} keep it discrete and autoregressive, next to continuous and discrete image diffusion respectively, while Dual Diffusion denoises text with masked discrete diffusion alongside a continuous image flow~\citep{li2025dualdiffusion}. UniDiffuser places both modalities in continuous latents and selects marginal, conditional or joint generation through their noise levels~\citep{bao2023unidiffuser}. OmniFlow extends this to rectified flows over image, text and audio with modality-specific flow times~\citep{li2025omniflow}, and JUICE trains a single flow-matching transformer over frozen image and text representations, again with a flow time per modality~\citep{li2026juice}. These flow times say where each stream lies between noise and data. Depth is a second, independent coordinate, which says how far each stream has been refined within one network evaluation.

Generating text with a flow requires a continuous representation that the generator can refine. LaDiC diffuses a continuous BERT latent conditioned on the image~\citep{wang2024ladic}, and OmniFlow compresses FLAN-T5 features into a text VAE latent~\citep{li2025omniflow}. Embedded Language Flows (ELF) avoid a separate latent and train directly on sequences of contextual text embeddings~\citep{elf2026}. MoLo follows this route, generating FLAN-T5 encoder features so that captioning and image synthesis share one endpoint-prediction objective, but reads them out with a frozen decoder, where ELF decodes with the denoising network itself.

\section{Conclusion}
\label{sec:conclusion}

We introduced Mixture of Looped Modalities (MoLo), which unties depth across modalities as Mixture-of-Transformers unties their weights. Each modality loops through its own core on a depth grid of its own, and because training samples these grids independently, a single checkpoint runs any allocation of image and text loops at inference, without retraining. In both directions, more loops improve the modality being generated, while the condition needs at most one loop. With 188M trainable parameters, the same checkpoint generates images and captions and has a higher prompt-following average than 13 of 15 published text-to-image generators.

Depth in a multimodal generator is therefore better treated as a setting chosen per modality, for the task and budget at hand, than as a fixed property of the architecture. Because loops add computation but no parameters, this also points toward compact multimodal models that adapt their compute to the device they run on. With two modalities, the condition already needs one loop as text and none as an image. With more modalities, or with several generated together, their needs are likely to diverge further, which is where depth chosen per modality should pay off most. Choosing the allocation per prompt or per integration step is a further direction.

\section*{Acknowledgements}

This project was supported by the ELLIS Unit Amsterdam and by SURF
through access to the Snellius supercomputer. Cees G. M. Snoek is (partially) funded by the Horizon Europe project ELLIOT (GA No. 101214398). Pedro M. P. Curvo is supported by the Dutch Research Council (NWO) through the OpenBots project (Grant No. 20442, DOI: 10.61686/TYGTC51084), under the Applied and Engineering Sciences (TTW) research programme.

\bibliographystyle{unsrtnat}
\bibliography{references}

\begin{thebibliography}{77}
\providecommand{\natexlab}[1]{#1}
\providecommand{\url}[1]{\texttt{#1}}
\expandafter\ifx\csname urlstyle\endcsname\relax
  \providecommand{\doi}[1]{doi: #1}\else
  \providecommand{\doi}{doi: \begingroup \urlstyle{rm}\Url}\fi

\bibitem[Vaswani et~al.(2017)Vaswani, Shazeer, Parmar, Uszkoreit, Jones, Gomez,
  Kaiser, and Polosukhin]{vaswani2017attention}
Ashish Vaswani, Noam Shazeer, Niki Parmar, Jakob Uszkoreit, Llion Jones,
  Aidan~N. Gomez, {\L}ukasz Kaiser, and Illia Polosukhin.
\newblock Attention is all you need.
\newblock In \emph{Advances in Neural Information Processing Systems},
  volume~30, 2017.
\newblock URL
  \url{https://proceedings.neurips.cc/paper/2017/hash/3f5ee243547dee91fbd053c1c4a845aa-Abstract.html}.

\bibitem[Peebles and Xie(2023)]{peebles2023}
William Peebles and Saining Xie.
\newblock Scalable diffusion models with transformers.
\newblock In \emph{Proceedings of the IEEE/CVF International Conference on
  Computer Vision}, pages 4195--4205, 2023.
\newblock URL
  \url{https://openaccess.thecvf.com/content/ICCV2023/html/Peebles_Scalable_Diffusion_Models_with_Transformers_ICCV_2023_paper.html}.

\bibitem[Esser et~al.(2024)Esser, Kulal, Blattmann, Entezari, M{\"u}ller,
  Saini, Levi, Lorenz, Sauer, Boesel, Podell, Dockhorn, English, and
  Rombach]{esser2024sd3}
Patrick Esser, Sumith Kulal, Andreas Blattmann, Rahim Entezari, Jonas
  M{\"u}ller, Harry Saini, Yam Levi, Dominik Lorenz, Axel Sauer, Frederic
  Boesel, Dustin Podell, Tim Dockhorn, Zion English, and Robin Rombach.
\newblock Scaling rectified flow transformers for high-resolution image
  synthesis.
\newblock In \emph{Proceedings of the 41st International Conference on Machine
  Learning}, volume 235 of \emph{Proceedings of Machine Learning Research},
  pages 12606--12633, 2024.
\newblock URL \url{https://proceedings.mlr.press/v235/esser24a.html}.

\bibitem[Polyak et~al.(2024)]{polyak2024moviegen}
Adam Polyak et~al.
\newblock {Movie Gen}: A cast of media foundation models.
\newblock \emph{arXiv preprint arXiv:2410.13720}, 2024.
\newblock URL \url{https://arxiv.org/abs/2410.13720}.

\bibitem[{NVIDIA}(2025)]{nvidia2025cosmos}
{NVIDIA}.
\newblock {Cosmos} world foundation model platform for physical {AI}.
\newblock \emph{arXiv preprint arXiv:2501.03575}, 2025.
\newblock URL \url{https://arxiv.org/abs/2501.03575}.

\bibitem[Hayes et~al.(2025)Hayes, Rao, Akin, Sofroniew, Oktay, Lin, Verkuil,
  Tran, Deaton, Wiggert, Badkundri, Shafkat, Gong, Derry, Molina, Thomas, Khan,
  Mishra, Kim, Bartie, Nemeth, Hsu, Sercu, Candido, and Rives]{hayes2025esm3}
Thomas Hayes, Roshan Rao, Halil Akin, Nicholas~J. Sofroniew, Deniz Oktay,
  Zeming Lin, Robert Verkuil, Vincent~Q. Tran, Jonathan Deaton, Marius Wiggert,
  Rohil Badkundri, Irhum Shafkat, Jun Gong, Alexander Derry, Raul~S. Molina,
  Neil Thomas, Yousuf~A. Khan, Chetan Mishra, Carolyn Kim, Liam~J. Bartie,
  Matthew Nemeth, Patrick~D. Hsu, Tom Sercu, Salvatore Candido, and Alexander
  Rives.
\newblock Simulating 500 million years of evolution with a language model.
\newblock \emph{Science}, 387\penalty0 (6736):\penalty0 850--858, 2025.
\newblock \doi{10.1126/science.ads0018}.
\newblock URL \url{https://doi.org/10.1126/science.ads0018}.

\bibitem[{Chameleon Team}(2024)]{chameleon2024}
{Chameleon Team}.
\newblock Chameleon: Mixed-modal early-fusion foundation models.
\newblock \emph{arXiv preprint arXiv:2405.09818}, 2024.
\newblock URL \url{https://arxiv.org/abs/2405.09818}.

\bibitem[Zhou et~al.(2025)Zhou, Yu, Babu, Tirumala, Yasunaga, Shamis, Kahn, Ma,
  Zettlemoyer, and Levy]{zhou2024transfusion}
Chunting Zhou, Lili Yu, Arun Babu, Kushal Tirumala, Michihiro Yasunaga, Leonid
  Shamis, Jacob Kahn, Xuezhe Ma, Luke Zettlemoyer, and Omer Levy.
\newblock Transfusion: Predict the next token and diffuse images with one
  multi-modal model.
\newblock In \emph{International Conference on Learning Representations}, 2025.
\newblock URL
  \url{https://proceedings.iclr.cc/paper_files/paper/2025/hash/12678c3948153f4bc391f51e2082bd6e-Abstract-Conference.html}.

\bibitem[Xie et~al.(2025{\natexlab{a}})Xie, Mao, Bai, Zhang, Wang, Lin, Gu,
  Chen, Yang, and Shou]{xie2024showo}
Jinheng Xie, Weijia Mao, Zechen Bai, David~Junhao Zhang, Weihao Wang,
  Kevin~Qinghong Lin, Yuchao Gu, Zhijie Chen, Zhenheng Yang, and Mike~Zheng
  Shou.
\newblock Show-o: One single transformer to unify multimodal understanding and
  generation.
\newblock In \emph{International Conference on Learning Representations},
  2025{\natexlab{a}}.
\newblock URL
  \url{https://proceedings.iclr.cc/paper_files/paper/2025/hash/45f0d179ef7e10eb7366550cd4e574ae-Abstract-Conference.html}.

\bibitem[Bao et~al.(2023)Bao, Nie, Xue, Li, Pu, Wang, Yue, Cao, Su, and
  Zhu]{bao2023unidiffuser}
Fan Bao, Shen Nie, Kaiwen Xue, Chongxuan Li, Shi Pu, Yaole Wang, Gang Yue, Yue
  Cao, Hang Su, and Jun Zhu.
\newblock One transformer fits all distributions in multi-modal diffusion at
  scale.
\newblock In \emph{Proceedings of the 40th International Conference on Machine
  Learning}, volume 202 of \emph{Proceedings of Machine Learning Research},
  pages 1692--1717, 2023.
\newblock URL \url{https://proceedings.mlr.press/v202/bao23a.html}.

\bibitem[Liang et~al.(2025)Liang, Yu, Luo, Iyer, Dong, Zhou, Ghosh, Lewis, Yih,
  Zettlemoyer, and Lin]{liang2024mot}
Weixin Liang, Lili Yu, Liang Luo, Srinivasan Iyer, Ning Dong, Chunting Zhou,
  Gargi Ghosh, Mike Lewis, Wen-tau Yih, Luke Zettlemoyer, and Xi~Victoria Lin.
\newblock Mixture-of-transformers: A sparse and scalable architecture for
  multi-modal foundation models.
\newblock \emph{Transactions on Machine Learning Research}, 2025.
\newblock URL \url{https://openreview.net/forum?id=Nu6N69i8SB}.

\bibitem[Dehghani et~al.(2019)Dehghani, Gouws, Vinyals, Uszkoreit, and
  Kaiser]{dehghani2019universal}
Mostafa Dehghani, Stephan Gouws, Oriol Vinyals, Jakob Uszkoreit, and {\L}ukasz
  Kaiser.
\newblock Universal transformers.
\newblock In \emph{International Conference on Learning Representations}, 2019.
\newblock URL \url{https://openreview.net/forum?id=HyzdRiR9Y7}.

\bibitem[Geiping et~al.(2025)Geiping, McLeish, Jain, Kirchenbauer, Singh,
  Bartoldson, Kailkhura, Bhatele, and Goldstein]{geiping2025recurrent}
Jonas Geiping, Sean McLeish, Neel Jain, John Kirchenbauer, Siddharth Singh,
  Brian Bartoldson, Bhavya Kailkhura, Abhinav Bhatele, and Tom Goldstein.
\newblock Scaling up test-time compute with latent reasoning: A recurrent depth
  approach.
\newblock In \emph{Advances in Neural Information Processing Systems},
  volume~38, pages 46242--46293, 2025.
\newblock \doi{10.52202/085713-1380}.
\newblock URL
  \url{https://proceedings.neurips.cc/paper_files/paper/2025/hash/3b01972cf31e6fa0fe29e4b8b5c2a0a1-Abstract-Conference.html}.

\bibitem[Lipman et~al.(2023)Lipman, Chen, Ben-Hamu, Nickel, and Le]{lipman2023}
Yaron Lipman, Ricky T.~Q. Chen, Heli Ben-Hamu, Maximilian Nickel, and Matt Le.
\newblock Flow matching for generative modeling.
\newblock In \emph{International Conference on Learning Representations}, 2023.
\newblock URL \url{https://openreview.net/forum?id=PqvMRDCJT9t}.

\bibitem[Liu et~al.(2023)Liu, Gong, and Liu]{liu2023rectified}
Xingchao Liu, Chengyue Gong, and Qiang Liu.
\newblock Flow straight and fast: Learning to generate and transfer data with
  rectified flow.
\newblock In \emph{International Conference on Learning Representations}, 2023.
\newblock URL \url{https://openreview.net/forum?id=XVjTT1nw5z}.

\bibitem[Albergo and Vanden-Eijnden(2023)]{albergo2023interpolants}
Michael~S. Albergo and Eric Vanden-Eijnden.
\newblock Building normalizing flows with stochastic interpolants.
\newblock In \emph{International Conference on Learning Representations}, 2023.
\newblock URL \url{https://openreview.net/forum?id=li7qeBbCR1t}.

\bibitem[Derakhshani et~al.(2026)Derakhshani, Curvo, Burghouts, van~de Meent,
  and Snoek]{derakhshani2026lift}
Mohammad~Mahdi Derakhshani, Pedro M.~P. Curvo, Gertjan~J. Burghouts, Jan-Willem
  van~de Meent, and Cees G.~M. Snoek.
\newblock {LiFT}: Loop flow transformers.
\newblock \emph{arXiv preprint arXiv:2610.05538}, 2026.
\newblock URL \url{https://arxiv.org/abs/2610.05538}.

\bibitem[Chng et~al.(2026)Chng, Chen, Tong, Diao, Cai, Yang, Liu, Lu, Lin, and
  Huang]{chng2026loopeddit}
Yong~Xien Chng, Tianyi Chen, Wenwen Tong, Haiwen Diao, Zhongang Cai, Lei Yang,
  Ziwei Liu, Lewei Lu, Dahua Lin, and Gao Huang.
\newblock Looped diffusion transformer.
\newblock arXiv:2609.40305, 2026.
\newblock URL \url{https://arxiv.org/abs/2609.40305}.

\bibitem[Chen et~al.(2018)Chen, Rubanova, Bettencourt, and
  Duvenaud]{chen2018neural}
Ricky T.~Q. Chen, Yulia Rubanova, Jesse Bettencourt, and David~K. Duvenaud.
\newblock Neural ordinary differential equations.
\newblock In \emph{Advances in Neural Information Processing Systems},
  volume~31, pages 6571--6583, 2018.
\newblock URL
  \url{https://proceedings.neurips.cc/paper_files/paper/2018/hash/69386f6bb1dfed68692a24c8686939b9-Abstract.html}.

\bibitem[Li and He(2026)]{li2026jit}
Tianhong Li and Kaiming He.
\newblock Back to basics: Let denoising generative models denoise.
\newblock In \emph{Proceedings of the IEEE/CVF Conference on Computer Vision
  and Pattern Recognition}, pages 36115--36125, 2026.
\newblock URL \url{https://arxiv.org/abs/2511.13720}.

\bibitem[Hu et~al.(2026)Hu, Qiu, Lu, Zhao, Li, Kim, Andreas, and He]{elf2026}
Keya Hu, Linlu Qiu, Yiyang Lu, Hanhong Zhao, Tianhong Li, Yoon Kim, Jacob
  Andreas, and Kaiming He.
\newblock {ELF}: Embedded language flows.
\newblock In \emph{Advances in Neural Information Processing Systems}, 2026.
\newblock URL \url{https://openreview.net/forum?id=z3I6Aj563f}.

\bibitem[Ho and Salimans(2022)]{ho2022classifierfree}
Jonathan Ho and Tim Salimans.
\newblock Classifier-free diffusion guidance.
\newblock \emph{arXiv preprint arXiv:2207.12598}, 2022.
\newblock URL \url{https://arxiv.org/abs/2207.12598}.

\bibitem[Chung et~al.(2024)Chung, Hou, Longpre, Zoph, Tay, Fedus, Li, Wang,
  Dehghani, Brahma, Webson, Gu, Dai, Suzgun, Chen, Chowdhery, Castro-Ros,
  Pellat, Robinson, Valter, Narang, Mishra, Yu, Zhao, Huang, Dai, Yu, Petrov,
  Chi, Dean, Devlin, Roberts, Zhou, Le, and Wei]{chung2024flant5}
Hyung~Won Chung, Le~Hou, Shayne Longpre, Barret Zoph, Yi~Tay, William Fedus,
  Yunxuan Li, Xuezhi Wang, Mostafa Dehghani, Siddhartha Brahma, Albert Webson,
  Shixiang~Shane Gu, Zhuyun Dai, Mirac Suzgun, Xinyun Chen, Aakanksha
  Chowdhery, Alex Castro-Ros, Marie Pellat, Kevin Robinson, Dasha Valter,
  Sharan Narang, Gaurav Mishra, Adams Yu, Vincent Zhao, Yanping Huang, Andrew
  Dai, Hongkun Yu, Slav Petrov, Ed~H. Chi, Jeff Dean, Jacob Devlin, Adam
  Roberts, Denny Zhou, Quoc~V. Le, and Jason Wei.
\newblock Scaling instruction-finetuned language models.
\newblock \emph{Journal of Machine Learning Research}, 25\penalty0
  (70):\penalty0 1--53, 2024.
\newblock URL \url{https://www.jmlr.org/papers/v25/23-0870.html}.

\bibitem[Changpinyo et~al.(2021)Changpinyo, Sharma, Ding, and
  Soricut]{changpinyo2021cc12m}
Soravit Changpinyo, Piyush Sharma, Nan Ding, and Radu Soricut.
\newblock Conceptual {12M}: Pushing web-scale image-text pre-training to
  recognize long-tail visual concepts.
\newblock In \emph{Proceedings of the IEEE/CVF Conference on Computer Vision
  and Pattern Recognition}, pages 3558--3568, 2021.
\newblock URL
  \url{https://openaccess.thecvf.com/content/CVPR2021/html/Changpinyo_Conceptual_12M_Pushing_Web-Scale_Image-Text_Pre-Training_To_Recognize_Long-Tail_Visual_CVPR_2021_paper.html}.

\bibitem[Fang et~al.(2026)Fang, Yu, Duan, Huang, Bai, Cai, Wang, Liu, Liu, and
  Li]{fang2026prism}
Rongyao Fang, Aldrich Yu, Chengqi Duan, Linjiang Huang, Shuai Bai, Yuxuan Cai,
  Kun Wang, Si~Liu, Xihui Liu, and Hongsheng Li.
\newblock {FLUX-Reason-6M} \& {PRISM-Bench}: A million-scale text-to-image
  reasoning dataset and comprehensive benchmark.
\newblock In \emph{International Conference on Learning Representations}, 2026.
\newblock URL
  \url{https://proceedings.iclr.cc/paper_files/paper/2026/hash/6724eae98f3917968d54c193ac0b45f1-Abstract-Conference.html}.

\bibitem[Lin et~al.(2014)Lin, Maire, Belongie, Hays, Perona, Ramanan,
  Doll{\'a}r, and Zitnick]{lin2014coco}
Tsung-Yi Lin, Michael Maire, Serge Belongie, James Hays, Pietro Perona, Deva
  Ramanan, Piotr Doll{\'a}r, and C.~Lawrence Zitnick.
\newblock Microsoft {COCO}: Common objects in context.
\newblock In \emph{European Conference on Computer Vision}, pages 740--755,
  2014.
\newblock \doi{10.1007/978-3-319-10602-1_48}.
\newblock URL
  \url{https://link.springer.com/chapter/10.1007/978-3-319-10602-1_48}.

\bibitem[Heusel et~al.(2017)Heusel, Ramsauer, Unterthiner, Nessler, and
  Hochreiter]{heusel2017fid}
Martin Heusel, Hubert Ramsauer, Thomas Unterthiner, Bernhard Nessler, and Sepp
  Hochreiter.
\newblock {GANs} trained by a two time-scale update rule converge to a local
  nash equilibrium.
\newblock In \emph{Advances in Neural Information Processing Systems},
  volume~30, 2017.
\newblock URL
  \url{https://proceedings.neurips.cc/paper/2017/hash/8a1d694707eb0fefe65871369074926d-Abstract.html}.

\bibitem[Jayasumana et~al.(2024)Jayasumana, Ramalingam, Veit, Glasner,
  Chakrabarti, and Kumar]{jayasumana2024cmmd}
Sadeep Jayasumana, Srikumar Ramalingam, Andreas Veit, Daniel Glasner, Ayan
  Chakrabarti, and Sanjiv Kumar.
\newblock Rethinking {FID}: Towards a better evaluation metric for image
  generation.
\newblock In \emph{Proceedings of the IEEE/CVF Conference on Computer Vision
  and Pattern Recognition}, pages 9307--9315, 2024.
\newblock URL
  \url{https://openaccess.thecvf.com/content/CVPR2024/papers/Jayasumana_Rethinking_FID_Towards_a_Better_Evaluation_Metric_for_Image_Generation_CVPR_2024_paper.pdf}.

\bibitem[Hessel et~al.(2021)Hessel, Holtzman, Forbes, Le~Bras, and
  Choi]{hessel2021clipscore}
Jack Hessel, Ari Holtzman, Maxwell Forbes, Ronan Le~Bras, and Yejin Choi.
\newblock {CLIPScore}: A reference-free evaluation metric for image captioning.
\newblock In \emph{Conference on Empirical Methods in Natural Language
  Processing}, pages 7514--7528, 2021.
\newblock URL \url{https://aclanthology.org/2021.emnlp-main.595/}.

\bibitem[Zhang et~al.(2024)Zhang, Zhang, Dong, Zang, and
  Wang]{zhang2024longclip}
Beichen Zhang, Pan Zhang, Xiaoyi Dong, Yuhang Zang, and Jiaqi Wang.
\newblock {Long-CLIP}: Unlocking the long-text capability of {CLIP}.
\newblock In \emph{European Conference on Computer Vision}, pages 310--325,
  2024.
\newblock \doi{10.1007/978-3-031-72983-6_18}.
\newblock URL
  \url{https://link.springer.com/chapter/10.1007/978-3-031-72983-6_18}.

\bibitem[Ghosh et~al.(2023)Ghosh, Hajishirzi, and Schmidt]{ghosh2023geneval}
Dhruba Ghosh, Hannaneh Hajishirzi, and Ludwig Schmidt.
\newblock {GenEval}: An object-focused framework for evaluating text-to-image
  alignment.
\newblock In \emph{Advances in Neural Information Processing Systems},
  volume~36, 2023.
\newblock URL
  \url{https://proceedings.neurips.cc/paper_files/paper/2023/hash/a3bf71c7c63f0c3bcb7ff67c67b1e7b1-Abstract-Datasets_and_Benchmarks.html}.

\bibitem[Hu et~al.(2024)Hu, Wang, Fang, Fu, Cheng, and Yu]{hu2024ella}
Xiwei Hu, Rui Wang, Yixiao Fang, Bin Fu, Pei Cheng, and Gang Yu.
\newblock {ELLA}: Equip diffusion models with {LLM} for enhanced semantic
  alignment.
\newblock arXiv:2403.05135, 2024.
\newblock URL \url{https://arxiv.org/abs/2403.05135}.

\bibitem[Li et~al.(2026{\natexlab{a}})Li, Wang, Hu, Huang, Chen, Ou, Tao, Wan,
  Qi, and Feng]{li2026corebench}
Ouxiang Li, Yuan Wang, Xinting Hu, Huijuan Huang, Rui Chen, Jiarong Ou, Xin
  Tao, Pengfei Wan, Xiaojuan Qi, and Fuli Feng.
\newblock Easier painting than thinking: Can text-to-image models set the
  stage, but not direct the play?
\newblock In \emph{International Conference on Learning Representations},
  2026{\natexlab{a}}.
\newblock URL
  \url{https://proceedings.iclr.cc/paper_files/paper/2026/hash/8bb0d291acd4acf06ef112099c16f326-Abstract-Conference.html}.

\bibitem[Wang et~al.(2026{\natexlab{a}})Wang, Hu, Wang, Xiong, Zhang, and
  Chu]{wang2026spatial}
Zengbin Wang, Xuecai Hu, Yong Wang, Feng Xiong, Man Zhang, and Xiangxiang Chu.
\newblock Everything in its place: Benchmarking spatial intelligence of
  text-to-image models.
\newblock In \emph{International Conference on Learning Representations},
  2026{\natexlab{a}}.
\newblock URL
  \url{https://proceedings.iclr.cc/paper_files/paper/2026/hash/4a5c76c63f83ea45fb136d62db6c7104-Abstract-Conference.html}.

\bibitem[Wang et~al.(2026{\natexlab{b}})Wang, Zhao, Lu, Zhou, Ma, and
  He]{wang2026minit2i}
Xianbang Wang, Hanhong Zhao, Yiyang Lu, Kangyang Zhou, Linrui Ma, and Kaiming
  He.
\newblock {MiniT2I}: A minimalist baseline for text-to-image generation.
\newblock Blog post, 2026{\natexlab{b}}.
\newblock URL \url{https://peppaking8.github.io/#/post/minit2i}.

\bibitem[Shen et~al.(2025)Shen, Yu, Zhou, Li, and Barsoum]{shen2025emmdit}
Tong Shen, Jingai Yu, Dong Zhou, Dong Li, and Emad Barsoum.
\newblock {E-MMDiT}: Revisiting multimodal diffusion transformer design for
  fast image synthesis under limited resources.
\newblock arXiv:2510.27135, 2025.
\newblock URL \url{https://arxiv.org/abs/2510.27135}.

\bibitem[Feng et~al.(2026)Feng, Wei, Chen, Pan, Liu, Yan, Gao, and
  Zuo]{feng2026dreamlite}
Kailai Feng, Yuxiang Wei, Bo~Chen, Yang Pan, Songwei Liu, Chenqian Yan, Yuan
  Gao, and Wangmeng Zuo.
\newblock {DreamLite}: A lightweight on-device unified model for image
  generation and editing.
\newblock In \emph{Computer Vision -- ECCV 2026}, volume 17067 of \emph{Lecture
  Notes in Computer Science}, pages 214--232, 2026.
\newblock \doi{10.1007/978-3-032-37174-4_13}.
\newblock URL
  \url{https://link.springer.com/chapter/10.1007/978-3-032-37174-4_13}.

\bibitem[Xie et~al.(2025{\natexlab{b}})Xie, Chen, Chen, Cai, Tang, Lin, Zhang,
  Li, Zhu, Lu, and Han]{xie2025sana}
Enze Xie, Junsong Chen, Junyu Chen, Han Cai, Haotian Tang, Yujun Lin, Zhekai
  Zhang, Muyang Li, Ligeng Zhu, Yao Lu, and Song Han.
\newblock {SANA}: Efficient high-resolution text-to-image synthesis with linear
  diffusion transformers.
\newblock In \emph{International Conference on Learning Representations},
  2025{\natexlab{b}}.
\newblock URL
  \url{https://proceedings.iclr.cc/paper_files/paper/2025/hash/34345e243156da67605d4b63d71c8d98-Abstract-Conference.html}.

\bibitem[Deng et~al.(2026)Deng, Pan, Zhang, Liu, Luo, Cui, Wang, Shen, Shan,
  Zhang, and Wang]{deng2025ursa}
Haoge Deng, Ting Pan, Fan Zhang, Yang Liu, Zhuoyan Luo, Yufeng Cui, Wenxuan
  Wang, Chunhua Shen, Shiguang Shan, Zhaoxiang Zhang, and Xinlong Wang.
\newblock Uniform discrete diffusion with metric path for video generation.
\newblock In \emph{International Conference on Learning Representations}, 2026.
\newblock URL
  \url{https://proceedings.iclr.cc/paper_files/paper/2026/hash/daf8364f0715a41a469c677c0adc4754-Abstract-Conference.html}.

\bibitem[Wang et~al.(2026{\natexlab{c}})Wang, Zhang, Yue, Yue, Li, Ouyang, and
  Bai]{wang2025transition}
Zidong Wang, Yiyuan Zhang, Xiaoyu Yue, Xiangyu Yue, Yangguang Li, Wanli Ouyang,
  and Lei Bai.
\newblock Transition models: Rethinking the generative learning objective.
\newblock In \emph{Proceedings of the IEEE/CVF Conference on Computer Vision
  and Pattern Recognition}, pages 29178--29189, 2026{\natexlab{c}}.
\newblock URL
  \url{https://openaccess.thecvf.com/content/CVPR2026/papers/Wang_Transition_Models_Rethinking_the_Generative_Learning_Objective_CVPR_2026_paper.pdf}.

\bibitem[Ma et~al.(2026{\natexlab{a}})Ma, Wei, Wang, Zhang, and
  Tian]{ma2025deco}
Zehong Ma, Longhui Wei, Shuai Wang, Shiliang Zhang, and Qi~Tian.
\newblock {DeCo}: Frequency-decoupled pixel diffusion for end-to-end image
  generation.
\newblock In \emph{Proceedings of the IEEE/CVF Conference on Computer Vision
  and Pattern Recognition}, pages 43600--43610, 2026{\natexlab{a}}.
\newblock URL
  \url{https://openaccess.thecvf.com/content/CVPR2026/html/Ma_DeCo_Frequency-Decoupled_Pixel_Diffusion_for_End-to-End_Image_Generation_CVPR_2026_paper.html}.

\bibitem[Tang et~al.(2026)Tang, Xie, Bao, Weng, Li, Zheng, and
  Wang]{tang2025unilip}
Hao Tang, Chen-Wei Xie, Xiaoyi Bao, Tingyu Weng, Pandeng Li, Yun Zheng, and
  Liwei Wang.
\newblock {UniLiP}: Adapting {CLIP} for unified multimodal understanding,
  generation and editing.
\newblock In \emph{International Conference on Learning Representations}, 2026.
\newblock URL
  \url{https://proceedings.iclr.cc/paper_files/paper/2026/hash/627e8695eca215f4268880f571193dc4-Abstract-Conference.html}.

\bibitem[Tian et~al.(2026)Tian, Yang, Chen, Cui, Wang, Duan, Yin, Chen, Yang,
  Liu, Zhu, Fan, Gu, Wang, Wei, Yin, Yang, Zhong, Qin, Xin, Fu, Liu, Ge, Guo,
  Luo, Li, Qiao, Chen, and Zhang]{tian2026internvlu}
Changyao Tian, Danni Yang, Guanzhou Chen, Erfei Cui, Zhaokai Wang, Yuchen Duan,
  Penghao Yin, Sitao Chen, Ganlin Yang, Mingxin Liu, Zirun Zhu, Ziqian Fan,
  Leyao Gu, Haomin Wang, Qi~Wei, Jinhui Yin, Xue Yang, Zhihang Zhong, Qi~Qin,
  Yi~Xin, Bin Fu, Yihao Liu, Jiaye Ge, Qipeng Guo, Gen Luo, Hongsheng Li,
  Yu~Qiao, Kai Chen, and Hongjie Zhang.
\newblock {InternVL-U}: Democratizing unified multimodal models for
  understanding, reasoning, generation and editing.
\newblock arXiv:2603.09877, 2026.
\newblock URL \url{https://arxiv.org/abs/2603.09877}.

\bibitem[{Zhipu AI}(2025)]{cogview42025}
{Zhipu AI}.
\newblock {CogView4-6B}.
\newblock Model card, 2025.
\newblock URL \url{https://huggingface.co/zai-org/CogView4-6B}.

\bibitem[Zheng et~al.(2024)Zheng, Teng, Yang, Wang, Chen, Gu, Dong, Ding, and
  Tang]{zheng2024cogview3}
Wendi Zheng, Jiayan Teng, Zhuoyi Yang, Weihan Wang, Jidong Chen, Xiaotao Gu,
  Yuxiao Dong, Ming Ding, and Jie Tang.
\newblock {CogView3}: Finer and faster text-to-image generation via relay
  diffusion.
\newblock In \emph{Computer Vision -- ECCV 2024}, volume 15135 of \emph{Lecture
  Notes in Computer Science}, pages 1--22, 2024.
\newblock \doi{10.1007/978-3-031-72980-5_1}.
\newblock URL
  \url{https://link.springer.com/chapter/10.1007/978-3-031-72980-5_1}.

\bibitem[Qin et~al.(2026)Qin, Gong, Sun, Li, Pan, Yang, Yang, Qu, Tan, and
  Li]{qin2025unicot}
Luozheng Qin, Jia Gong, Yuqing Sun, Tianjiao Li, Haoyu Pan, Mengping Yang,
  Xiaomeng Yang, Chao Qu, Zhiyu Tan, and Hao Li.
\newblock {Uni-CoT}: Towards unified chain-of-thought reasoning across text and
  vision.
\newblock In \emph{International Conference on Learning Representations}, 2026.
\newblock URL
  \url{https://proceedings.iclr.cc/paper_files/paper/2026/hash/1ae4999aefb509d75d8608e07280922c-Abstract-Conference.html}.

\bibitem[Duan et~al.(2026)Duan, Fang, Wang, Wang, Huang, Zeng, Li, and
  Liu]{duan2025gotr1}
Chengqi Duan, Rongyao Fang, Yuqing Wang, Kun Wang, Linjiang Huang, Xingyu Zeng,
  Hongsheng Li, and Xihui Liu.
\newblock {GoT-R1}: Unleashing reasoning capability of autoregressive visual
  generation with reinforcement learning.
\newblock In \emph{International Conference on Learning Representations}, 2026.
\newblock URL
  \url{https://proceedings.iclr.cc/paper_files/paper/2026/hash/c0c165157df7e3082f9c6d70d3a4b6e9-Abstract-Conference.html}.

\bibitem[Jiang et~al.(2025)Jiang, Guo, Zhang, Zong, Li, Zhuo, Yan, Heng, and
  Li]{jiang2025t2ir1}
Dongzhi Jiang, Ziyu Guo, Renrui Zhang, Zhuofan Zong, Hao Li, Le~Zhuo, Shilin
  Yan, Pheng-Ann Heng, and Hongsheng Li.
\newblock {T2I-R1}: Reinforcing image generation with collaborative
  semantic-level and token-level {CoT}.
\newblock In \emph{Advances in Neural Information Processing Systems},
  volume~38, pages 44581--44615, 2025.
\newblock \doi{10.52202/085713-1330}.
\newblock URL
  \url{https://proceedings.neurips.cc/paper_files/paper/2025/hash/38fc6254f73450813db3b3e04397a9fc-Abstract-Conference.html}.

\bibitem[Lin et~al.(2024)Lin, Shrivastava, Luo, Iyer, Lewis, Ghosh,
  Zettlemoyer, and Aghajanyan]{lin2024moma}
Xi~Victoria Lin, Akshat Shrivastava, Liang Luo, Srinivasan Iyer, Mike Lewis,
  Gargi Ghosh, Luke Zettlemoyer, and Armen Aghajanyan.
\newblock {MoMa}: Efficient early-fusion pre-training with mixture of
  modality-aware experts.
\newblock \emph{arXiv preprint arXiv:2407.21770}, 2024.
\newblock URL \url{https://arxiv.org/abs/2407.21770}.

\bibitem[Saharia et~al.(2022)Saharia, Chan, Saxena, Li, Whang, Denton,
  Ghasemipour, Gontijo~Lopes, Karagol~Ayan, Salimans, Ho, Fleet, and
  Norouzi]{saharia2022imagen}
Chitwan Saharia, William Chan, Saurabh Saxena, Lala Li, Jay Whang, Emily~L.
  Denton, Kamyar Ghasemipour, Raphael Gontijo~Lopes, Burcu Karagol~Ayan, Tim
  Salimans, Jonathan Ho, David~J. Fleet, and Mohammad Norouzi.
\newblock Photorealistic text-to-image diffusion models with deep language
  understanding.
\newblock In \emph{Advances in Neural Information Processing Systems},
  volume~35, pages 36479--36494, 2022.
\newblock \doi{10.52202/068431-2643}.
\newblock URL
  \url{https://proceedings.neurips.cc/paper_files/paper/2022/hash/ec795aeadae0b7d230fa35cbaf04c041-Abstract-Conference.html}.

\bibitem[Chen et~al.(2024)Chen, Yu, Ge, Yao, Xie, Wu, Wang, Kwok, Luo, Lu, and
  Li]{chen2024pixartalpha}
Junsong Chen, Jincheng Yu, Chongjian Ge, Lewei Yao, Enze Xie, Yue Wu, Zhongdao
  Wang, James Kwok, Ping Luo, Huchuan Lu, and Zhenguo Li.
\newblock {PixArt-$\alpha$}: Fast training of diffusion transformer for
  photorealistic text-to-image synthesis.
\newblock In \emph{International Conference on Learning Representations}, 2024.
\newblock URL
  \url{https://proceedings.iclr.cc/paper_files/paper/2024/hash/fe989bb038b5dcc44181255dd6913e43-Abstract-Conference.html}.

\bibitem[Schwarzschild et~al.(2021)Schwarzschild, Borgnia, Gupta, Huang,
  Vishkin, Goldblum, and Goldstein]{schwarzschild2021algorithm}
Avi Schwarzschild, Eitan Borgnia, Arjun Gupta, Furong Huang, Uzi Vishkin, Micah
  Goldblum, and Tom Goldstein.
\newblock Can you learn an algorithm? generalizing from easy to hard problems
  with recurrent networks.
\newblock In \emph{Advances in Neural Information Processing Systems},
  volume~34, pages 6695--6706, 2021.
\newblock URL
  \url{https://proceedings.neurips.cc/paper_files/paper/2021/hash/3501672ebc68a5524629080e3ef60aef-Abstract.html}.

\bibitem[Bansal et~al.(2022)Bansal, Schwarzschild, Borgnia, Emam, Huang,
  Goldblum, and Goldstein]{bansal2022extrapolation}
Arpit Bansal, Avi Schwarzschild, Eitan Borgnia, Zeyad Emam, Furong Huang, Micah
  Goldblum, and Tom Goldstein.
\newblock End-to-end algorithm synthesis with recurrent networks: Extrapolation
  without overthinking.
\newblock In \emph{Advances in Neural Information Processing Systems},
  volume~35, pages 20232--20242, 2022.
\newblock \doi{10.52202/068431-1471}.
\newblock URL
  \url{https://proceedings.neurips.cc/paper_files/paper/2022/hash/7f70331dbe58ad59d83941dfa7d975aa-Abstract-Conference.html}.

\bibitem[Xu and Sato(2025)]{xu2025timestep}
Kevin Xu and Issei Sato.
\newblock On expressive power of looped transformers: Theoretical analysis and
  enhancement via timestep encoding.
\newblock In \emph{Proceedings of the 42nd International Conference on Machine
  Learning}, volume 267 of \emph{Proceedings of Machine Learning Research},
  pages 69613--69646, 2025.
\newblock URL \url{https://proceedings.mlr.press/v267/xu25x.html}.

\bibitem[Jeddi et~al.(2026)Jeddi, Ciccone, and Taati]{jeddi2026loopformer}
Ahmadreza Jeddi, Marco Ciccone, and Babak Taati.
\newblock {LoopFormer}: Elastic-depth looped transformers for latent reasoning
  via shortcut modulation.
\newblock In \emph{International Conference on Learning Representations}, 2026.
\newblock URL
  \url{https://proceedings.iclr.cc/paper_files/paper/2026/hash/4edfc7fb472789e31de455f8735ecbc3-Abstract-Conference.html}.

\bibitem[Raposo et~al.(2024)Raposo, Ritter, Richards, Lillicrap, Humphreys, and
  Santoro]{raposo2024mod}
David Raposo, Sam Ritter, Blake Richards, Timothy Lillicrap, Peter~Conway
  Humphreys, and Adam Santoro.
\newblock Mixture-of-depths: Dynamically allocating compute in
  transformer-based language models.
\newblock \emph{arXiv preprint arXiv:2404.02258}, 2024.
\newblock URL \url{https://arxiv.org/abs/2404.02258}.

\bibitem[Luo et~al.(2025)Luo, Luo, Ji, Zhou, Sun, Shen, and
  Ji]{luo2025gammamod}
Yaxin Luo, Gen Luo, Jiayi Ji, Yiyi Zhou, Xiaoshuai Sun, Zhiqiang Shen, and
  Rongrong Ji.
\newblock {$\gamma$-MoD}: Exploring mixture-of-depth adaptation for multimodal
  large language models.
\newblock In \emph{International Conference on Learning Representations}, 2025.
\newblock URL
  \url{https://proceedings.iclr.cc/paper_files/paper/2025/hash/b3847cda0c8cc0cfcdacf462dc122214-Abstract-Conference.html}.

\bibitem[Zhang et~al.(2025)Zhang, Meng, Zhang, Huang, Wu, and
  Wang]{zhang2025pmod}
Jun Zhang, Desen Meng, Zhengming Zhang, Zhenpeng Huang, Tao Wu, and Limin Wang.
\newblock {p-MoD}: Building mixture-of-depths {MLLMs} via progressive ratio
  decay.
\newblock In \emph{Proceedings of the IEEE/CVF International Conference on
  Computer Vision}, pages 3705--3715, 2025.
\newblock URL
  \url{https://openaccess.thecvf.com/content/ICCV2025/html/Zhang_p-MoD_Building_Mixture-of-Depths_MLLMs_via_Progressive_Ratio_Decay_ICCV_2025_paper.html}.

\bibitem[Bae et~al.(2025)Bae, Kim, Bayat, Kim, Ha, Schuster, Fisch,
  Harutyunyan, Ji, Courville, and Yun]{bae2025mor}
Sangmin Bae, Yujin Kim, Reza Bayat, Sungnyun Kim, Jiyoun Ha, Tal Schuster, Adam
  Fisch, Hrayr Harutyunyan, Ziwei Ji, Aaron~C. Courville, and Se-Young Yun.
\newblock Mixture-of-recursions: Learning dynamic recursive depths for adaptive
  token-level computation.
\newblock In \emph{Advances in Neural Information Processing Systems},
  volume~38, pages 107088--107133, 2025.
\newblock \doi{10.52202/085713-3229}.
\newblock URL
  \url{https://proceedings.neurips.cc/paper_files/paper/2025/hash/8b08bbf8b420faa6eeb4020720582ec7-Abstract-Conference.html}.

\bibitem[Xu et~al.(2026)Xu, Gao, Wang, Li, Chen, Guo, Yang, and
  Zhang]{xu2026recursivevlm}
Ruihan Xu, Yuting Gao, Lan Wang, Jianing Li, Weihao Chen, Qingpei Guo, Ming
  Yang, and Shiliang Zhang.
\newblock Looping back to move forward: Recursive transformers for efficient
  and flexible large multimodal models.
\newblock \emph{arXiv preprint arXiv:2602.09080}, 2026.
\newblock URL \url{https://arxiv.org/abs/2602.09080}.

\bibitem[Chen et~al.(2025{\natexlab{a}})Chen, Qian, Hu, Fu, Tong, Wang, Li,
  Zhang, Schwing, Liu, and Yang]{chen2025ditair}
Chen Chen, Rui Qian, Wenze Hu, Tsu-Jui Fu, Jialing Tong, Xinze Wang, Lezhi Li,
  Bowen Zhang, Alex Schwing, Wei Liu, and Yinfei Yang.
\newblock {DiT-Air}: Revisiting the efficiency of diffusion model architecture
  design in text to image generation.
\newblock \emph{arXiv preprint arXiv:2503.10618}, 2025{\natexlab{a}}.
\newblock URL \url{https://arxiv.org/abs/2503.10618}.

\bibitem[Chai(2026)]{chai2026loopdit}
Wenhao Chai.
\newblock {LoopDiT}: Loop transformers for diffusion models.
\newblock Technical blog post, 2026.
\newblock URL \url{https://wenhaochai.com/exps/loopdit.html}.

\bibitem[Shin et~al.(2025)Shin, Yang, and Chen]{shin2025deepflow}
Inkyu Shin, Chenglin Yang, and Liang-Chieh Chen.
\newblock Deeply supervised flow-based generative models.
\newblock In \emph{Proceedings of the IEEE/CVF International Conference on
  Computer Vision}, pages 16535--16544, 2025.
\newblock URL
  \url{https://openaccess.thecvf.com/content/ICCV2025/html/Shin_Deeply_Supervised_Flow-Based_Generative_Models_ICCV_2025_paper.html}.

\bibitem[Goyal et~al.(2026)Goyal, Agrawal, Anil, Jain, Paul, and
  Kusupati]{goyal2026elt}
Sahil Goyal, Swayam Agrawal, Gautham~Govind Anil, Prateek Jain, Sujoy Paul, and
  Aditya Kusupati.
\newblock {ELT}: Elastic looped transformers for visual generation.
\newblock In \emph{Computer Vision -- ECCV 2026}, volume 17056 of \emph{Lecture
  Notes in Computer Science}, pages 308--326, 2026.
\newblock \doi{10.1007/978-3-032-37324-3_17}.
\newblock URL
  \url{https://link.springer.com/chapter/10.1007/978-3-032-37324-3_17}.

\bibitem[Wang and An(2026)]{wang2026distilltoloop}
Chenru Wang and Shengwei An.
\newblock Distill to loop: Turning pretrained diffusion transformers into
  looped models with adjustable depth.
\newblock \emph{arXiv preprint arXiv:2611.31977}, 2026.
\newblock URL \url{https://arxiv.org/abs/2611.31977}.

\bibitem[Jabri et~al.(2023)Jabri, Fleet, and Chen]{jabri2023rin}
Allan Jabri, David~J. Fleet, and Ting Chen.
\newblock Scalable adaptive computation for iterative generation.
\newblock In \emph{Proceedings of the 40th International Conference on Machine
  Learning}, volume 202 of \emph{Proceedings of Machine Learning Research},
  pages 14569--14589, 2023.
\newblock URL \url{https://proceedings.mlr.press/v202/jabri23a.html}.

\bibitem[Suleymanzade et~al.(2026)Suleymanzade, Lee, Eijkelboom, Boffi, Ceylan,
  and Kim]{suleymanzade2026thinking}
Ayhan Suleymanzade, Chanhyuk Lee, Floor Eijkelboom, Nicholas~M. Boffi,
  {\.{I}}smail~{\.{I}}lkan Ceylan, and Jinwoo Kim.
\newblock Thinking with looped flows.
\newblock \emph{arXiv preprint arXiv:2609.11801}, 2026.
\newblock URL \url{https://arxiv.org/abs/2609.11801}.

\bibitem[Li et~al.(2025{\natexlab{a}})Li, Li, Shi, Farimani, Kluger, Yang, and
  Wang]{li2025dualdiffusion}
Zijie Li, Henry Li, Yichun Shi, Amir~Barati Farimani, Yuval Kluger, Linjie
  Yang, and Peng Wang.
\newblock Dual diffusion for unified image generation and understanding.
\newblock In \emph{Proceedings of the IEEE/CVF Conference on Computer Vision
  and Pattern Recognition}, pages 2779--2790, 2025{\natexlab{a}}.
\newblock URL
  \url{https://openaccess.thecvf.com/content/CVPR2025/html/Li_Dual_Diffusion_for_Unified_Image_Generation_and_Understanding_CVPR_2025_paper.html}.

\bibitem[Li et~al.(2025{\natexlab{b}})Li, Kallidromitis, Gokul, Liao, Kato,
  Kozuka, and Grover]{li2025omniflow}
Shufan Li, Konstantinos Kallidromitis, Akash Gokul, Zichun Liao, Yusuke Kato,
  Kazuki Kozuka, and Aditya Grover.
\newblock {OmniFlow}: Any-to-any generation with multi-modal rectified flows.
\newblock In \emph{Proceedings of the IEEE/CVF Conference on Computer Vision
  and Pattern Recognition}, pages 13178--13188, 2025{\natexlab{b}}.
\newblock URL
  \url{https://openaccess.thecvf.com/content/CVPR2025/html/Li_OmniFlow_Any-to-Any_Generation_with_Multi-Modal_Rectified_Flows_CVPR_2025_paper.html}.

\bibitem[Li et~al.(2026{\natexlab{b}})Li, Su, Tong, Yang, Guo, Huang, Zhao,
  Sharma, Yang, Gu, Fergus, and Cho]{li2026juice}
Jiahan Li, Huangyuan Su, Shengbang Tong, Shusheng Yang, Ruihan Guo, Suozhi
  Huang, Hexu Zhao, Pratyusha Sharma, Sherry Yang, Jiatao Gu, Rob Fergus, and
  Kyunghyun Cho.
\newblock {JUICE}: Unified multimodal modeling in pretrained representation
  spaces.
\newblock arXiv:2610.28801, 2026{\natexlab{b}}.
\newblock URL \url{https://arxiv.org/abs/2610.28801}.

\bibitem[Wang et~al.(2024)Wang, Ren, Gao, Yao, Guo, An, Bai, and
  Sun]{wang2024ladic}
Yuchi Wang, Shuhuai Ren, Rundong Gao, Linli Yao, Qingyan Guo, Kaikai An,
  Jianhong Bai, and Xu~Sun.
\newblock {LaDiC}: Are diffusion models really inferior to autoregressive
  counterparts for image-to-text generation?
\newblock In \emph{Proceedings of the 2024 Conference of the North American
  Chapter of the Association for Computational Linguistics: Human Language
  Technologies}, pages 6699--6715, 2024.
\newblock \doi{10.18653/v1/2024.naacl-long.373}.
\newblock URL \url{https://aclanthology.org/2024.naacl-long.373/}.

\bibitem[{Caption Emporium}(2024)]{captionemporium2024cc12m}
{Caption Emporium}.
\newblock Conceptual {12M} with {LLaVA-NeXT} captions.
\newblock Hugging Face dataset, 2024.
\newblock URL
  \url{https://huggingface.co/datasets/CaptionEmporium/conceptual-captions-cc12m-llavanext}.

\bibitem[Liu et~al.(2024)Liu, Li, Li, Li, Zhang, Shen, and
  Lee]{liu2024llavanext}
Haotian Liu, Chunyuan Li, Yuheng Li, Bo~Li, Yuanhan Zhang, Sheng Shen, and
  Yong~Jae Lee.
\newblock {LLaVA-NeXT}: Improved reasoning, {OCR}, and world knowledge.
\newblock Blog post, 2024.
\newblock URL \url{https://llava-vl.github.io/blog/2024-01-30-llava-next/}.

\bibitem[Chen et~al.(2025{\natexlab{b}})Chen, Xu, Pan, Hu, Qin, Goldstein,
  Huang, Zhou, Xie, Savarese, Xue, Xiong, and Xu]{chen2025blip3o}
Jiuhai Chen, Zhiyang Xu, Xichen Pan, Yushi Hu, Can Qin, Tom Goldstein, Lifu
  Huang, Tianyi Zhou, Saining Xie, Silvio Savarese, Le~Xue, Caiming Xiong, and
  Ran Xu.
\newblock {BLIP3-o}: A family of fully open unified multimodal
  models---architecture, training and dataset.
\newblock \emph{arXiv preprint arXiv:2505.09568}, 2025{\natexlab{b}}.
\newblock URL \url{https://arxiv.org/abs/2505.09568}.

\bibitem[Nagengast et~al.(2024)Nagengast, Pach, Maltsev, Egan, and {LAION
  Community}]{nagengast2024dalle3discord}
Zach Nagengast, Eduardo Pach, Seva Maltsev, Ben Egan, and {LAION Community}.
\newblock {LAION DALL-E 3} discord dataset.
\newblock Hugging Face dataset, 2024.
\newblock URL
  \url{https://huggingface.co/datasets/OpenDatasets/dalle-3-dataset}.

\bibitem[Chen et~al.(2025{\natexlab{c}})Chen, Cai, Chen, Chen, Ji, Wang, Yang,
  and Wang]{chen2025sharegpt4oimg}
Junying Chen, Zhenyang Cai, Pengcheng Chen, Shunian Chen, Ke~Ji, Xidong Wang,
  Yunjin Yang, and Benyou Wang.
\newblock {ShareGPT-4o-Image}: Aligning multimodal models with {GPT-4o}-level
  image generation.
\newblock \emph{arXiv preprint arXiv:2506.18095}, 2025{\natexlab{c}}.
\newblock URL \url{https://arxiv.org/abs/2506.18095}.

\bibitem[Ma et~al.(2026{\natexlab{b}})Ma, Zhang, Dong, and Fu]{ma2026finet2i}
Xu~Ma, Yitian Zhang, Qihua Dong, and Yun Fu.
\newblock {Fine-T2I}: An open, large-scale, and diverse dataset for
  high-quality {T2I} fine-tuning.
\newblock \emph{arXiv preprint arXiv:2602.09439}, 2026{\natexlab{b}}.
\newblock URL \url{https://arxiv.org/abs/2602.09439}.

\end{thebibliography}

\clearpage
\appendix
\section*{Appendix index}
\pdfbookmark[0]{Appendix index}{appendix-index}

\begingroup
\small
\setlength{\tabcolsep}{0pt}
\renewcommand{\arraystretch}{1.25}
\newcommand{\appentry}[2]{%
  \hyperref[#1]{\ref*{#1}\quad #2} & \pageref{#1}\\}
\newcommand{\appgroup}[2]{%
  \multicolumn{2}{@{}l@{}}{\textbf{\hyperref[#1]{\ref*{#1}\quad #2}}}\\[2pt]}
\begin{tabular}{@{}p{.93\linewidth}r@{}}
 & Page\\
\appgroup{sec:app-experimental-details}{Experimental Details}
\appentry{sec:app-architecture}{Architecture}
\appentry{sec:app-pseudocode}{One Network Evaluation}
\appentry{sec:app-training}{Data, Flow and Optimization}
\appentry{sec:app-evaluation}{Evaluation Protocol}
\appentry{sec:app-cc12m-scaling}{The CC12M Scaling Campaign}
\noalign{\vskip 8pt}
\appgroup{sec:app-results}{Additional Results}
\appentry{sec:benchmark-allocation-details}{Prompt Following Across Loop Allocations}
\appentry{sec:benchmark-categories}{Benchmark Scores by Category}
\appentry{sec:app-frequency}{Frequency Analysis Across Loops}
\appentry{sec:caption-results}{Caption Results}
\end{tabular}
\endgroup

\clearpage
\section{Experimental Details}
\label{sec:app-experimental-details}
\begingroup\small

\subsection{Architecture}
\label{sec:app-architecture}
The generator has 187,791,104 trainable parameters, counting each recurrent weight once and excluding the frozen FLAN encoder and caption decoder. We project $16\times16$ patches of the $3\times512\times512$ image and the $256\times1{,}024$ caption features to width 768. Both streams use 12 attention heads and a common rotary encoding with axes for text position, image row and image column. Each block applies attention, with normalized queries and keys, and a SwiGLU feed-forward branch of hidden width 2,048 in parallel, with RMS normalization and stage-shared, modality-specific modulation. At the start of each active pass, the stream adds its post-fusion state to its RMS-normalized current state. Four private prelude blocks, two joint MMDiT fusion blocks~\citep{esser2024sd3} and four passes through the six-block core give 30 block applications per stream at training depth. The generated stream has a direct prediction head. Because the image holds four times as many tokens as the caption, an image-only event costs nearly three times the generator FLOPs of a text-only event (\cref{tab:cost}).

We truncate or pad captions to 256 positions and run the frozen FP32 FLAN encoder under BF16 autocast. For image synthesis, observed tokens supply their raw final encoder states and padded positions receive a learned null text vector. For captioning, all 256 raw states, including those at padded positions, serve as generation targets. The frozen decoder reads all 256 generated states and decodes greedily without the reference caption length. It is a FLAN-T5-Large decoder trained separately for 20,000 updates to read these states, independently of the generator, and every reported caption score uses it.

\begin{table}[!htbp]
\centering\small
\caption{\textbf{An image loop costs nearly three times the generator compute of a text loop.} Generator TFLOPs and end-to-end H100 latency at batch size one, per image at 100 integration steps and guidance scale 6, and per caption at 32 integration steps and guidance scale 2. Rows give image loops and columns text loops. Caption latency includes the frozen decoder and is the median of 20 timings after three warm-up runs.}
\label{tab:cost}
\begin{tabular}{@{}l*{6}{r}@{}}
\toprule
& \multicolumn{6}{c}{Text loops} \\
\cmidrule(l){2-7}
Image loops & 0 & 1 & 2 & 4 & 8 & 16 \\
\midrule
\multicolumn{7}{@{}l}{\emph{Text to image, TFLOPs per image}} \\
1 & 50.3 & 58.7 & 63.5 & 80.5 & 114.3 & 182.0 \\
2 & 73.5 & 81.9 & 83.1 & 100.0 & 133.9 & 201.5 \\
4 & 119.8 & 128.3 & 129.5 & 139.2 & 173.0 & 240.7 \\
8 & 212.6 & 221.1 & 222.3 & 232.0 & 251.3 & 318.9 \\
16 & 398.2 & 406.6 & 407.9 & 417.5 & 436.8 & 475.5 \\
\midrule
\multicolumn{7}{@{}l}{\emph{Text to image, seconds per image}} \\
1 & 4.02 & 5.62 & 6.03 & 9.23 & 15.49 & 27.56 \\
2 & 5.67 & 7.22 & 6.67 & 9.66 & 15.94 & 28.30 \\
4 & 8.99 & 10.56 & 9.90 & 10.74 & 17.07 & 29.57 \\
8 & 15.40 & 17.50 & 16.93 & 17.57 & 19.35 & 31.65 \\
16 & 28.95 & 30.10 & 29.63 & 30.52 & 32.30 & 35.39 \\
\midrule
\multicolumn{7}{@{}l}{\emph{Image to text, TFLOPs per caption}} \\
0 & -- & 11.3 & 14.1 & 19.5 & 30.5 & 52.3 \\
1 & -- & 18.7 & 21.4 & 26.9 & 37.8 & 59.7 \\
2 & -- & 24.9 & 26.4 & 31.9 & 42.8 & 64.7 \\
4 & -- & 39.6 & 41.1 & 44.3 & 55.2 & 77.1 \\
8 & -- & 68.9 & 70.5 & 73.7 & 79.9 & 101.8 \\
16 & -- & 127.7 & 129.3 & 132.4 & 138.7 & 151.3 \\
\midrule
\multicolumn{7}{@{}l}{\emph{Image to text, median seconds per caption}} \\
0 & -- & 1.99 & 2.67 & 3.78 & 5.69 & 9.67 \\
1 & -- & 2.44 & 3.14 & 4.12 & 6.29 & 10.27 \\
2 & -- & 2.66 & 2.93 & 3.90 & 6.12 & 10.06 \\
4 & -- & 3.64 & 3.98 & 4.16 & 6.34 & 10.23 \\
8 & -- & 5.63 & 6.09 & 6.19 & 6.90 & 10.82 \\
16 & -- & 9.65 & 10.07 & 10.43 & 10.86 & 12.34 \\
\bottomrule
\end{tabular}
\end{table}

\subsection{One Network Evaluation}
\label{sec:app-pseudocode}
\Cref{alg:molo-evaluation} gives the computation of one network evaluation for a set of streams $\mathcal M$ with generated stream $g$ (\cref{sec:molo}). Training and inference run the same procedure and differ only in the grids. During training, each stream draws its grid independently and the trajectory loss of \cref{eq:molo-trajectory} supervises every estimate $y_k$. At inference, the allocation fixes the grids (\cref{sec:molo-inference}), and the sampler converts the final estimate $y_{K_g}$, made at depth $1$, into a velocity for the next integration step. A waiting stream keeps its latest state, from which every core block recomputes its keys and values; gradients flow through these projections and the state. At inference these keys and values stay fixed while the stream waits, so caching them would give identical outputs at lower cost, and the FLOP counts in \cref{tab:cost} include the recomputation.

\begin{algorithm}[htbp]
\caption{One MoLo network evaluation with a depth grid per stream}
\label{alg:molo-evaluation}
\begin{algorithmic}[1]
\REQUIRE Streams $\mathcal M$ with inputs $x_m$ at generative times $t_m$, generated stream $g$, task $\tau$, grids $\mathcal S_m\subset(0,1]$ with $\max\mathcal S_g=1$, core blocks $\ell=1,\ldots,L$
\STATE $(h^0_m)_{m\in\mathcal M}\gets F\big((P_m(x_m;t,\tau))_{m\in\mathcal M}\big)$ \COMMENT{Private preludes, then joint fusion}
\STATE $h_m\gets h^0_m$ and $\delta_m\gets0$ for every $m\in\mathcal M$ \COMMENT{Latest state and the depth it reached}
\STATE $y_0\gets D_g(h^0_g)$; $k\gets0$ \COMMENT{Initial estimate; $\operatorname{sg}(y_0)$ anchors the targets}
\STATE Sort the distinct depths of $\bigcup_{m}\mathcal S_m$ into $\sigma_1<\cdots<\sigma_J$
\FOR{$j=1,\ldots,J$}
  \STATE $\mathcal A_j\gets\{m\in\mathcal M:\sigma_j\in\mathcal S_m\}$ \COMMENT{Active streams}
  \STATE $u_m\gets\operatorname{Norm}(h_m)+h^0_m$ and $\delta_m\gets\sigma_j$ for every $m\in\mathcal A_j$ \COMMENT{Reinject the initial state}
  \FOR{$\ell=1,\ldots,L$}
    \STATE Every $m\in\mathcal M$ projects keys and values from $u_m$ if $m\in\mathcal A_j$ and from $h_m$ otherwise, conditioned on $\delta_m$, $t$ and $\tau$
    \STATE Every $m\in\mathcal A_j$ forms queries, attends over all keys and values as in \cref{eq:bg-joint-attention}, and updates $u_m$ with its own block-$\ell$ weights
  \ENDFOR
  \STATE $h_m\gets u_m$ for every $m\in\mathcal A_j$
  \IF{$g\in\mathcal A_j$}
    \STATE $k\gets k+1$; $y_k\gets D_g(h_g)$ \COMMENT{Estimate at depth $\sigma_j$}
  \ENDIF
\ENDFOR
\RETURN $y_0,y_1,\ldots,y_{K_g}$ \COMMENT{$y_{K_g}$ is the endpoint prediction at depth $1$}
\end{algorithmic}
\end{algorithm}

\subsection{Data, Flow and Optimization}
\label{sec:app-training}
Pretraining samples CC12M with LLaVA-NeXT captions~\citep{captionemporium2024cc12m,liu2024llavanext} and FLUX-Reason~\citep{fang2026prism} with i1 captions in proportions of approximately 65\% and 35\%. During fine-tuning the relative sampling weights for BLIP3-o-60K~\citep{chen2025blip3o}, DALL-E~3 Discord~\citep{nagengast2024dalle3discord}, ShareGPT-4o-Image~\citep{chen2025sharegpt4oimg} and Fine-T2I~\citep{ma2026finet2i} are $60:16:40:116$. We stream paired RGB images and captions in both stages. For image synthesis we sample the image flow time $t_I$ and keep the caption clean at $t_T=1$. For captioning we sample $t_T$ and keep the image clean at $t_I=1$. The generated stream is supervised on its clean endpoint in pixel or encoder-feature space.

Images are scaled to $[-1,1]$, and image noise has standard deviation 2 in RGB space and text noise has standard deviation 0.156 in encoder-feature space. We sample the outer time as $t=\operatorname{sigmoid}(z)$, where $z$ has standard deviation 0.8 and mean $-0.8$ for images or $-1.5$ for text. Conditioning dropout is 10\% in each direction, enabling classifier-free guidance for both tasks. A dropped condition is replaced by a learned null, and at guidance scale $\omega$ the sampler combines the conditional and unconditional endpoint predictions as $y_u+\omega\,(y_c-y_u)$, so that scale 1 disables guidance. The endpoint loss in both directions uses $w(t)=1/\max(1-t,0.05)^2$. The two directions carry equal weight, and the initial-prediction penalty has coefficient $\lambda=0.01$ in both.

For each training batch we draw three independent uniform depths in $(0,1)$ for each stream, sort them within each stream and append one. The two four-pass schedules are sampled independently, so the intermediate passes can interleave in either order and the two final passes coincide.

Optimization uses AdamW with $\beta=(0.9,0.95)$, zero weight decay and gradient clipping at 0.1. The global batch is 1,024, with eight examples per GPU microbatch and BF16 mixed precision. The learning rate warms up linearly from $10^{-6}$ to $4\times10^{-4}$ over 1,000 updates and then stays constant. The fine-tuning stage carries forward the optimizer state and the exponential moving average of the parameters, whose decay is 0.99995.

\subsection{Evaluation Protocol}
\label{sec:app-evaluation}
We score text-to-image generation on five prompt-following benchmarks. GenEval~\citep{ghosh2023geneval} checks objects, counts, colors and positions with the official detector. DPG-Bench~\citep{hu2024ella} tests whether multiple details of a dense prompt appear in the image, PRISM~\citep{fang2026prism} covers a broad range of reasoning and rendering tasks, and SpatialGenEval~\citep{wang2026spatial} probes spatial relationships. T2I-CoReBench~\citep{li2026corebench} scores each image against a prompt-specific checklist with a visual judge, and its official protocol left 92 judgments unresolved across our allocations, against about 54,000 per allocation. Only these were asked again, with the judge reasoning before it answers for the overall score and its final word read directly for the group scores. Counting every unresolved judgment as a failure, or every one as a pass, changes no allocation's score by more than 0.03 points. The overall score averages all checklist items, so it differs from the mean of the twelve group scores.

We evaluate the final EMA weights and keep prompts and random seeds fixed across loop allocations. In the allocation sweeps we vary the generated stream over $\{1,2,4,8,16\}$ loops and the conditioning stream over $\{0,1,2,4,8,16\}$, giving 30 allocations per direction. At each integration step we place the $K_g$ passes of the generated stream at depths $k/K_g$ for $k=1,\dots,K_g$. With two or more loops, the conditioning stream uses its own uniform grid. A single conditioning loop takes its pass at $1/(2K_g)$, halfway to the first generated pass, and with no conditioning loops the generated stream reads the conditioning state from the initial fusion throughout recurrence. Inference timings include the sampler and online encoding or text decoding, exclude disk I/O and first-use compilation, and use synchronized H100 measurements. FLOP estimates count the generator's matrix multiplications. CMMD follows the reference implementation of \citet{jayasumana2024cmmd}, with CLIP ViT-L/14 embeddings at 336 pixels and values scaled by $10^3$. Captions are scored with Long-CLIP-B~\citep{zhang2024longclip}, which reads 248 tokens where CLIPScore reads 77. Each caption receives the prefix \emph{A photo depicts}, which counts toward that window together with the start and end tokens, and the score averages $2.5\max(0,\cos(v,t))$ over photographs, where $v$ and $t$ are the image and caption embeddings. On 5,000 COCO photographs, its 95\% bootstrap interval is about $\pm0.0016$.

\subsection{The CC12M Scaling Campaign}
\label{sec:app-cc12m-scaling}

The scaling campaign uses a common bank of CC12M prompts, reference images and generation seeds. The CC12M bank is drawn from the pretraining distribution. Every allocation is scored on the same prompts, references and seeds, so this shared advantage does not favor any of them. Each of three grids evaluates $K_I\in\{1,2,4,8,16\}$, crossed with $K_T\in\{0,1,2,4,8,16\}$ without guidance at 100 integration steps, with guidance scales from 1 to 6 at $K_T=1$ and 100 steps, or with 10, 25, 50 and 100 steps at $K_T=1$ without guidance. The grids have 30, 30 and 20 cells and share 10 repeated entries, for 70 distinct settings. The scores are available for nested banks of 1,000, 2,000, 5,000 and 10,000 samples. Figures in the main text use the 10,000-sample bank for every setting, and \cref{tab:cc12m-grid} lists its scores.

The two distribution metrics respond differently to the sample count, so we compare settings only within one bank at equal sample counts. At eight image loops and one text loop, with 100 integration steps and guidance disabled, FID falls from 20.22 to 13.56 between 5,000 and 10,000 samples, whereas CMMD changes by less than 0.001, consistent with the sample-size sensitivity of FID documented by \citet[Sec.~6.3]{jayasumana2024cmmd}. Because every setting is scored once on one bank, differences of a few tenths of an FID point, such as the 0.15 between eight and sixteen image loops, cannot be ranked reliably.

\begin{table}[!htbp]
\centering\footnotesize
\setlength{\tabcolsep}{2.6pt}
\caption{\textbf{One text loop gives the lowest FID and CMMD at every image depth.} FID and CMMD on the 10,000-sample CC12M bank for the three grids of \cref{sec:app-cc12m-scaling}, which also underlie \cref{fig:cc12m-text,fig:cc12m-cfg,fig:cc12m-steps}. Rows give image loops, and bold marks the best value in each row of each block. Guidance scale 1 disables guidance.}
\label{tab:cc12m-grid}
\begin{tabular}{@{}lrrrrrrrrrrrr@{}}
\toprule
\multicolumn{13}{@{}l}{\emph{Text loops $K_T$, without guidance, 100 steps}} \\
& \multicolumn{6}{c}{FID $\downarrow$} & \multicolumn{6}{c}{CMMD $\downarrow$} \\
\cmidrule(lr){2-7}\cmidrule(lr){8-13}
$K_I$ & 0 & 1 & 2 & 4 & 8 & 16 & 0 & 1 & 2 & 4 & 8 & 16 \\
1 & 38.17 & \textbf{31.55} & 46.62 & 46.73 & 46.80 & 46.91 & 2.044 & \textbf{1.760} & 2.200 & 2.200 & 2.202 & 2.203 \\
2 & 19.38 & \textbf{17.83} & 20.26 & 20.33 & 20.38 & 20.43 & 1.402 & \textbf{1.164} & 1.453 & 1.448 & 1.450 & 1.451 \\
4 & 15.38 & \textbf{14.34} & 17.96 & 15.63 & 15.76 & 15.81 & 1.113 & \textbf{0.891} & 1.230 & 1.142 & 1.147 & 1.150 \\
8 & 14.57 & \textbf{13.56} & 15.90 & 16.19 & 14.67 & 14.76 & 0.995 & \textbf{0.774} & 1.068 & 1.127 & 1.029 & 1.029 \\
16 & 14.54 & \textbf{13.71} & 15.68 & 16.11 & 16.25 & 14.63 & 0.968 & \textbf{0.751} & 1.017 & 1.070 & 1.129 & 1.015 \\
\midrule
\multicolumn{13}{@{}l}{\emph{Guidance scale, $K_T=1$, 100 steps}} \\
& \multicolumn{6}{c}{FID $\downarrow$} & \multicolumn{6}{c}{CMMD $\downarrow$} \\
\cmidrule(lr){2-7}\cmidrule(lr){8-13}
$K_I$ & 1 & 2 & 3 & 4 & 5 & 6 & 1 & 2 & 3 & 4 & 5 & 6 \\
1 & 31.55 & 15.71 & 14.51 & \textbf{14.46} & 14.58 & 14.71 & 1.760 & 1.062 & 0.843 & 0.748 & 0.702 & \textbf{0.679} \\
2 & 17.83 & \textbf{13.66} & 14.46 & 15.03 & 15.23 & 15.45 & 1.164 & 0.631 & 0.529 & 0.496 & 0.482 & \textbf{0.481} \\
4 & 14.34 & \textbf{14.05} & 15.00 & 15.30 & 15.45 & 15.61 & 0.891 & 0.533 & 0.476 & 0.457 & \textbf{0.450} & \textbf{0.450} \\
8 & \textbf{13.56} & 14.72 & 15.34 & 15.75 & 15.81 & 15.93 & 0.774 & 0.518 & 0.476 & 0.463 & \textbf{0.457} & 0.459 \\
16 & \textbf{13.71} & 14.92 & 15.51 & 15.76 & 15.81 & 15.87 & 0.751 & 0.519 & 0.481 & 0.465 & \textbf{0.462} & 0.464 \\
\midrule
\multicolumn{13}{@{}l}{\emph{Integration steps, $K_T=1$, without guidance}} \\
& \multicolumn{4}{c}{FID $\downarrow$} & \multicolumn{4}{c}{CMMD $\downarrow$} & & & & \\
\cmidrule(lr){2-5}\cmidrule(lr){6-9}
$K_I$ & 10 & 25 & 50 & 100 & 10 & 25 & 50 & 100 & & & & \\
1 & 41.30 & 33.82 & 32.56 & \textbf{31.55} & 1.882 & 1.832 & 1.780 & \textbf{1.760} & & & & \\
2 & 24.46 & 19.08 & 18.47 & \textbf{17.83} & 1.353 & 1.231 & 1.192 & \textbf{1.164} & & & & \\
4 & 20.29 & 14.97 & 14.61 & \textbf{14.34} & 1.135 & 0.948 & 0.914 & \textbf{0.891} & & & & \\
8 & 17.07 & \textbf{13.23} & 13.47 & 13.56 & 0.985 & 0.804 & 0.787 & \textbf{0.774} & & & & \\
16 & 16.36 & \textbf{13.15} & 13.55 & 13.71 & 0.938 & 0.772 & 0.764 & \textbf{0.751} & & & & \\
\bottomrule
\end{tabular}
\end{table}

\par\endgroup

\clearpage
\section{Additional Results}
\label{sec:app-results}
The results in this section complete those of \cref{sec:q1-depth,sec:q3-budget,sec:q4-comparison} on the metrics the main text uses. Loop pairs are ordered (image, text), and dashes denote unmeasured settings.

\subsection{Prompt Following Across Loop Allocations}
\label{sec:benchmark-allocation-details}

\begin{figure}[!htb]\centering
\includegraphics[width=\linewidth]{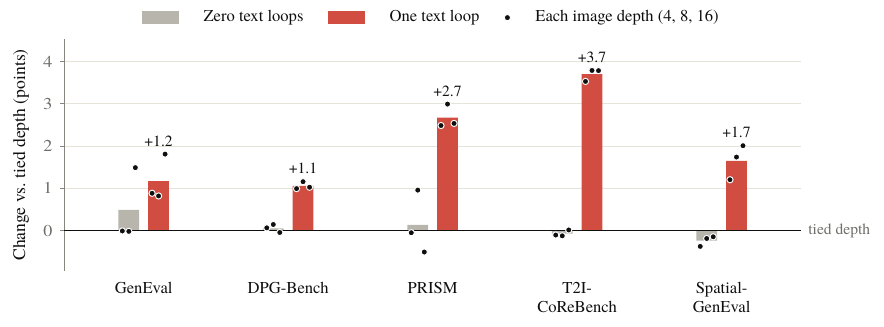}
\caption{\textbf{One early text loop improves every benchmark over tied depth, while zero text loops stay close to it.} Score change relative to tied depth ($K_T=K_I$) on five prompt-following benchmarks, at 100 integration steps and guidance scale 6. Bars give the mean over four, eight and sixteen image loops, with the value above each one-text-loop bar, and dots mark the individual image depths. \Cref{tab:text-loop-benchmarks} lists every score.}
\label{fig:text-loop-benchmarks}
\end{figure}

One early text loop scores above both zero text loops and tied depth on all five benchmarks at four, eight and sixteen image loops (\cref{fig:text-loop-benchmarks,tab:text-loop-benchmarks}), and it also beats the training allocation: $(4,1)$ improves every benchmark over $(4,4)$ at lower generator compute (\cref{tab:image-allocation}). The ordering extends below four image loops, where one text loop again scores highest at each image depth on the four benchmarks measured there (\cref{tab:benchmark-depth-extension}). Integration steps shift the best image depth instead: with 25 or 50 steps, GenEval peaks at four image loops rather than two (\cref{tab:benchmark-step-extension}).

\begin{table}[!htb]\centering\small
\caption{\textbf{One early text loop improves every benchmark over the training allocation, at lower compute.} The training allocation $(4,4)$ is compared with one early text loop at four and sixteen image loops. The last column gives the best value over all evaluated allocations, fourteen for every benchmark except T2I-CoReBench, which has nine. Generator compute is per image at 100 integration steps and guidance scale 6. Higher scores are better.}\label{tab:image-allocation}
\begin{tabular}{@{}lrrrl@{}}\toprule
& Training $(4,4)$ & $(4,1)$ & $(16,1)$ & Best at $(K_I,K_T)$ \\ \midrule
GenEval & 84.57 & 85.46 & 85.12 & 85.64 at $(2,1)$ \\
DPG-Bench & 83.52 & 84.51 & 84.58 & 84.80 at $(2,1)$ \\
PRISM & 56.45 & 58.94 & 58.69 & 59.09 at $(8,1)$ \\
T2I-CoReBench & 45.71 & 49.24 & 49.27 & 49.34 at $(8,1)$ \\
SpatialGenEval & 50.63 & 51.84 & 52.18 & 52.68 at $(2,1)$ \\ \midrule
Generator TFLOPs per image & 139.2 & 128.3 & 406.6 & \\ \bottomrule
\end{tabular}
\end{table}

\begin{table}[!htbp]
\centering\small
\setlength{\tabcolsep}{3.5pt}
\caption{\textbf{One text loop scores highest at every image depth on every benchmark.} Prompt-following scores at four, eight and sixteen image loops with zero text loops (0), one early text loop (1) or tied depth ($K_T=K_I$, =), at 100 integration steps and guidance scale 6. Bold marks the best of the three at each depth. \Cref{fig:text-loop-benchmarks} summarizes these scores relative to tied depth.}
\label{tab:text-loop-benchmarks}
\begin{tabular}{@{}l*{9}{r}@{}}
\toprule
& \multicolumn{3}{c}{4 image loops} & \multicolumn{3}{c}{8 image loops} & \multicolumn{3}{c}{16 image loops} \\
\cmidrule(lr){2-4}\cmidrule(lr){5-7}\cmidrule(l){8-10}
Text loops & 0 & 1 & = & 0 & 1 & = & 0 & 1 & = \\
\midrule
GenEval & 84.56 & \textbf{85.46} & 84.57 & 84.33 & \textbf{85.17} & 84.35 & 84.79 & \textbf{85.12} & 83.31 \\
DPG-Bench & 83.58 & \textbf{84.51} & 83.52 & 83.57 & \textbf{84.58} & 83.43 & 83.50 & \textbf{84.58} & 83.55 \\
PRISM & 56.40 & \textbf{58.94} & 56.45 & 57.06 & \textbf{59.09} & 56.10 & 55.64 & \textbf{58.69} & 56.15 \\
T2I-CoReBench & 45.60 & \textbf{49.24} & 45.71 & 45.42 & \textbf{49.34} & 45.55 & 45.50 & \textbf{49.27} & 45.49 \\
SpatialGenEval & 50.26 & \textbf{51.84} & 50.63 & 50.18 & \textbf{52.11} & 50.37 & 50.02 & \textbf{52.18} & 50.17 \\
\bottomrule
\end{tabular}
\end{table}

\begin{table}[htbp]
\centering
\caption{Low-depth prompt following at 100 integration steps. Column headings give $(K_I,K_T)$. Higher scores are better, and bold marks the best setting within each row.}
\label{tab:benchmark-depth-extension}
\begin{tabular}{@{}lrrrrr@{}}
\toprule
Benchmark & $(1,0)$ & $(1,1)$ & $(2,0)$ & $(2,1)$ & $(2,2)$ \\
\midrule
GenEval & 82.11 & 82.49 & 85.12 & \textbf{85.64} & 84.83 \\
DPG-Bench & 82.53 & 84.04 & 83.51 & \textbf{84.80} & 83.77 \\
PRISM & 53.89 & 55.69 & 56.79 & \textbf{59.09} & 55.85 \\
SpatialGenEval & 49.89 & 52.28 & 50.63 & \textbf{52.68} & 50.57 \\
\bottomrule
\end{tabular}
\end{table}

\begin{table}[htbp]
\centering
\caption{GenEval across integration steps and image depth, with one text loop. The 100-step row uses archived reference scores. Bold marks the best image depth for each step count.}
\label{tab:benchmark-step-extension}
\begin{tabular}{@{}lrrrr@{}}
\toprule
Steps & 2 image loops & 4 image loops & 8 image loops & 16 image loops \\
\midrule
25 & 82.39 & \textbf{83.10} & 82.99 & 82.04 \\
50 & 84.65 & \textbf{85.01} & 84.48 & 83.90 \\
100 & \textbf{85.64} & 85.46 & 85.17 & 85.12 \\
\bottomrule
\end{tabular}
\end{table}

\FloatBarrier
\subsection{Benchmark Scores by Category}
\label{sec:benchmark-categories}

On T2I-CoReBench and PRISM, text rendering gains most from one early text loop. Over tied depth, the text-rendering group of T2I-CoReBench (C-TR) more than doubles at every image depth (\cref{tab:corebench-categories}), and that of PRISM gains six to seven points at eight and sixteen image loops, more than any other group (\cref{tab:prism-categories}). \Cref{tab:geneval-categories,tab:dpg-categories,tab:spatial-categories} break down the other three benchmarks at the same allocations.

\begin{table}[!htbp]
\centering\small
\setlength{\tabcolsep}{3.5pt}
\caption{\textbf{One early text loop scores highest on counting at every image depth.} GenEval pass rate per category, with the official score in the first row. Columns as in \cref{tab:text-loop-benchmarks}: zero text loops (0), one early text loop (1) or tied depth (=) at each image depth. Bold marks the best of the three.}
\label{tab:geneval-categories}
\begin{tabular}{@{}l*{9}{r}@{}}
\toprule
& \multicolumn{3}{c}{4 image loops} & \multicolumn{3}{c}{8 image loops} & \multicolumn{3}{c}{16 image loops} \\
\cmidrule(lr){2-4}\cmidrule(lr){5-7}\cmidrule(l){8-10}
Text loops & 0 & 1 & = & 0 & 1 & = & 0 & 1 & = \\
\midrule
Overall & 84.6 & \textbf{85.5} & 84.6 & 84.3 & \textbf{85.2} & 84.3 & 84.8 & \textbf{85.1} & 83.3 \\
\midrule
Single object & 99.1 & \textbf{100.0} & 99.4 & 99.1 & \textbf{100.0} & 99.4 & 99.4 & \textbf{100.0} & 99.7 \\
Two objects & 94.4 & \textbf{94.9} & \textbf{94.9} & 94.2 & \textbf{95.7} & 94.2 & 95.2 & \textbf{95.5} & 93.7 \\
Counting & 72.5 & \textbf{74.1} & 70.6 & 70.3 & \textbf{74.4} & 71.3 & 71.3 & \textbf{72.8} & 68.8 \\
Colors & 89.6 & 91.2 & \textbf{91.5} & 89.9 & 90.2 & \textbf{91.8} & 90.7 & 90.4 & \textbf{91.0} \\
Position & 80.0 & \textbf{81.3} & \textbf{81.3} & 79.5 & \textbf{80.8} & 80.5 & 78.3 & \textbf{80.3} & \textbf{80.3} \\
Color attribution & \textbf{71.8} & 71.3 & 69.8 & \textbf{73.0} & 70.0 & 69.0 & \textbf{74.0} & 71.8 & 66.5 \\
\bottomrule
\end{tabular}
\end{table}

\begin{table}[!htbp]
\centering\small
\setlength{\tabcolsep}{3.5pt}
\caption{\textbf{DPG-Bench by category.} Broad categories with their subcategories indented, and the overall score in the first row. Columns as in \cref{tab:text-loop-benchmarks}: zero text loops (0), one early text loop (1) or tied depth (=) at each image depth. Bold marks the best of the three.}
\label{tab:dpg-categories}
\begin{tabular}{@{}l*{9}{r}@{}}
\toprule
& \multicolumn{3}{c}{4 image loops} & \multicolumn{3}{c}{8 image loops} & \multicolumn{3}{c}{16 image loops} \\
\cmidrule(lr){2-4}\cmidrule(lr){5-7}\cmidrule(l){8-10}
Text loops & 0 & 1 & = & 0 & 1 & = & 0 & 1 & = \\
\midrule
Overall & 83.6 & \textbf{84.5} & 83.5 & 83.6 & \textbf{84.6} & 83.4 & 83.5 & \textbf{84.6} & 83.5 \\
\midrule
Global & 81.2 & 81.8 & \textbf{82.1} & 82.1 & \textbf{82.7} & 81.5 & 81.8 & \textbf{83.6} & 80.9 \\
Entity & 90.0 & \textbf{90.7} & 90.3 & 90.2 & \textbf{90.7} & 90.3 & 90.5 & \textbf{90.8} & 90.2 \\
\quad part & 86.3 & 86.5 & \textbf{87.1} & 85.5 & 86.1 & \textbf{86.7} & 86.9 & 85.0 & \textbf{87.1} \\
\quad state & 83.2 & \textbf{83.3} & 83.1 & \textbf{83.5} & 83.2 & 82.4 & \textbf{83.7} & 83.2 & 82.0 \\
\quad whole & 91.7 & \textbf{92.7} & 92.1 & 92.0 & \textbf{92.7} & 92.2 & 92.2 & \textbf{92.9} & 92.1 \\
Attribute & \textbf{87.9} & 87.8 & 87.4 & 87.8 & \textbf{88.0} & 87.5 & 87.7 & \textbf{88.2} & 87.8 \\
\quad color & \textbf{91.7} & 91.5 & 91.5 & 91.8 & \textbf{92.1} & 91.8 & 91.6 & \textbf{91.9} & 91.6 \\
\quad other & \textbf{84.9} & \textbf{84.9} & 84.2 & 84.3 & \textbf{85.0} & 84.3 & 84.6 & \textbf{85.5} & 84.7 \\
\quad shape & \textbf{80.3} & 79.0 & 78.6 & \textbf{79.5} & 78.6 & 77.7 & \textbf{78.6} & \textbf{78.6} & 78.2 \\
\quad size & \textbf{73.1} & \textbf{73.1} & 71.9 & 72.3 & \textbf{74.0} & 71.5 & 71.5 & 71.9 & \textbf{73.6} \\
\quad texture & 88.1 & \textbf{88.3} & 87.8 & \textbf{88.4} & 88.1 & 87.7 & 88.3 & \textbf{88.9} & 88.4 \\
Relation & 93.6 & \textbf{94.1} & \textbf{94.1} & 93.6 & \textbf{93.9} & \textbf{93.9} & 93.7 & 93.9 & \textbf{94.1} \\
\quad non-spatial & 88.3 & 89.1 & \textbf{89.8} & \textbf{89.8} & 86.1 & \textbf{89.8} & 88.3 & 89.1 & \textbf{89.8} \\
\quad spatial & 93.9 & \textbf{94.4} & \textbf{94.4} & 93.9 & \textbf{94.3} & 94.1 & 94.0 & 94.2 & \textbf{94.3} \\
Other & 86.0 & \textbf{87.2} & 86.4 & 84.4 & 86.8 & \textbf{87.6} & 84.8 & \textbf{88.8} & 87.2 \\
\quad count & 85.4 & 85.4 & \textbf{85.9} & 83.9 & 84.9 & \textbf{86.9} & 84.4 & \textbf{87.4} & 85.4 \\
\quad text & 90.0 & \textbf{96.0} & 90.0 & 88.0 & \textbf{96.0} & 92.0 & 88.0 & \textbf{96.0} & \textbf{96.0} \\
\bottomrule
\end{tabular}
\end{table}

\begin{table}[!htbp]
\centering\small
\setlength{\tabcolsep}{3.5pt}
\caption{\textbf{One early text loop scores highest on both alignment and aesthetics at every image depth.} PRISM alignment and aesthetics scores, then the score of each prompt group, with the overall score in the first row. Columns as in \cref{tab:text-loop-benchmarks}: zero text loops (0), one early text loop (1) or tied depth (=) at each image depth. Bold marks the best of the three.}
\label{tab:prism-categories}
\begin{tabular}{@{}l*{9}{r}@{}}
\toprule
& \multicolumn{3}{c}{4 image loops} & \multicolumn{3}{c}{8 image loops} & \multicolumn{3}{c}{16 image loops} \\
\cmidrule(lr){2-4}\cmidrule(lr){5-7}\cmidrule(l){8-10}
Text loops & 0 & 1 & = & 0 & 1 & = & 0 & 1 & = \\
\midrule
Overall & 56.4 & \textbf{58.9} & 56.5 & 57.1 & \textbf{59.1} & 56.1 & 55.6 & \textbf{58.7} & 56.2 \\
\midrule
Alignment & 62.4 & \textbf{64.4} & 62.1 & 62.4 & \textbf{64.7} & 61.7 & 61.8 & \textbf{64.3} & 62.2 \\
Aesthetics & 50.4 & \textbf{53.5} & 50.8 & 51.7 & \textbf{53.5} & 50.5 & 49.5 & \textbf{53.1} & 50.1 \\
\midrule
Affection & 69.1 & \textbf{72.4} & 71.4 & 71.5 & \textbf{73.8} & 71.3 & 70.0 & \textbf{72.1} & 69.9 \\
Composition & 74.3 & \textbf{77.7} & 74.4 & 74.7 & \textbf{75.2} & 72.7 & 71.7 & \textbf{75.2} & 74.0 \\
Entity & 48.0 & \textbf{51.2} & 49.6 & 47.5 & \textbf{50.8} & 49.6 & 46.7 & \textbf{50.8} & 49.0 \\
Imagination & 54.3 & \textbf{54.4} & 51.4 & \textbf{55.2} & 55.1 & 52.9 & 54.5 & \textbf{54.9} & 53.6 \\
Long text & 51.9 & \textbf{54.0} & 50.3 & 50.1 & \textbf{53.6} & 51.1 & 49.7 & \textbf{51.7} & 50.6 \\
Style & 72.1 & \textbf{73.3} & 71.3 & \textbf{73.9} & 73.3 & 70.4 & 71.5 & \textbf{74.6} & 70.4 \\
Text rendering & 25.4 & \textbf{29.7} & 27.0 & 26.7 & \textbf{32.0} & 24.9 & 25.6 & \textbf{31.7} & 25.6 \\
\bottomrule
\end{tabular}
\end{table}

\begin{table}[!htbp]
\centering\small
\setlength{\tabcolsep}{3.5pt}
\caption{\textbf{SpatialGenEval by scene.} Score for each scene, with the overall score in the first row. Columns as in \cref{tab:text-loop-benchmarks}: zero text loops (0), one early text loop (1) or tied depth (=) at each image depth. Bold marks the best of the three.}
\label{tab:spatial-categories}
\begin{tabular}{@{}l*{9}{r}@{}}
\toprule
& \multicolumn{3}{c}{4 image loops} & \multicolumn{3}{c}{8 image loops} & \multicolumn{3}{c}{16 image loops} \\
\cmidrule(lr){2-4}\cmidrule(lr){5-7}\cmidrule(l){8-10}
Text loops & 0 & 1 & = & 0 & 1 & = & 0 & 1 & = \\
\midrule
Overall & 50.3 & \textbf{51.8} & 50.6 & 50.2 & \textbf{52.1} & 50.4 & 50.0 & \textbf{52.2} & 50.2 \\
\midrule
Advertisement Design & \textbf{50.4} & 46.2 & 47.0 & \textbf{48.0} & 46.2 & 46.0 & 47.0 & \textbf{47.6} & 47.4 \\
Airport & 52.8 & \textbf{53.8} & 51.2 & 53.0 & \textbf{53.4} & 50.6 & 52.6 & \textbf{53.4} & 52.6 \\
Art Gallery & 44.2 & \textbf{48.2} & 47.2 & 43.2 & \textbf{48.0} & 42.0 & 44.6 & \textbf{48.4} & 44.2 \\
Beach & \textbf{57.4} & 56.8 & 56.0 & \textbf{58.6} & 57.4 & 55.4 & 55.8 & 57.2 & \textbf{58.0} \\
Cafe & 60.6 & \textbf{60.8} & 60.6 & 60.0 & \textbf{60.2} & 59.8 & 59.0 & \textbf{62.6} & 61.0 \\
Cartoon Design & 46.6 & \textbf{51.8} & 46.6 & 49.0 & 48.2 & \textbf{50.4} & 48.0 & \textbf{49.8} & 47.4 \\
Cityspace & 41.4 & \textbf{45.6} & 44.2 & 39.8 & \textbf{44.8} & 42.2 & 42.4 & \textbf{46.2} & 41.6 \\
Classroom & 56.2 & \textbf{57.2} & 55.2 & 54.2 & \textbf{56.0} & 53.0 & 55.6 & \textbf{56.0} & 55.2 \\
Desert & 48.6 & \textbf{51.0} & 48.2 & 47.4 & \textbf{49.0} & 47.0 & 47.6 & \textbf{48.6} & 48.0 \\
Forest & 52.8 & \textbf{54.4} & 50.8 & 52.0 & \textbf{55.4} & 53.2 & 52.4 & \textbf{53.6} & 50.6 \\
Human Activities & 47.2 & 46.0 & \textbf{47.4} & 45.0 & \textbf{50.4} & 48.8 & 47.0 & \textbf{49.4} & 48.4 \\
Kitchen & 60.0 & 60.2 & \textbf{60.4} & \textbf{60.2} & \textbf{60.2} & 58.8 & 57.8 & \textbf{61.6} & 60.6 \\
Library & 48.2 & \textbf{49.4} & 47.6 & 48.6 & 49.2 & \textbf{50.0} & 48.0 & \textbf{50.0} & 48.8 \\
Living Room & 52.0 & \textbf{54.6} & 52.4 & 53.0 & 52.6 & \textbf{53.4} & 51.6 & \textbf{55.0} & 51.2 \\
Mountain & 52.8 & \textbf{54.0} & 53.6 & 52.0 & 54.6 & \textbf{55.0} & 50.0 & \textbf{53.4} & 51.4 \\
Office & 42.8 & \textbf{44.2} & 43.8 & 40.6 & \textbf{51.6} & 42.2 & 42.8 & \textbf{47.4} & 41.4 \\
Park & 59.2 & \textbf{60.2} & 56.4 & 59.8 & 58.0 & \textbf{60.0} & 58.0 & \textbf{59.8} & 57.8 \\
Portrait & 50.0 & 51.8 & \textbf{52.0} & 49.8 & \textbf{54.0} & 50.0 & 49.0 & \textbf{53.2} & 49.6 \\
Railway Station & 42.2 & \textbf{44.4} & 42.6 & 43.2 & 45.2 & \textbf{45.8} & 44.4 & \textbf{44.8} & 44.0 \\
Shopping Mall & 46.9 & 51.0 & \textbf{51.8} & 49.6 & \textbf{53.5} & 49.8 & 49.8 & \textbf{50.6} & 49.6 \\
Sports & 51.0 & \textbf{51.6} & 49.6 & 50.4 & \textbf{51.2} & 50.0 & 51.0 & \textbf{52.8} & 49.6 \\
Story Design & 51.0 & 51.7 & \textbf{54.0} & 51.7 & \textbf{52.3} & \textbf{52.3} & 52.0 & 51.7 & \textbf{52.7} \\
Underwater & 42.6 & 46.6 & \textbf{47.2} & 44.6 & \textbf{48.6} & 45.8 & 46.0 & \textbf{46.4} & 44.4 \\
Village & 49.0 & \textbf{50.8} & 50.2 & 49.2 & \textbf{52.2} & 48.8 & 49.8 & \textbf{51.6} & 49.4 \\
Zoo & 50.8 & \textbf{53.6} & 51.2 & \textbf{52.2} & 50.6 & 49.6 & 49.2 & \textbf{53.2} & 50.4 \\
\bottomrule
\end{tabular}
\end{table}

\begin{table}[!htbp]
\centering\small
\setlength{\tabcolsep}{3.5pt}
\caption{\textbf{T2I-CoReBench by dimension.} Group codes follow the benchmark. Composition (C) covers multi-instance (MI), multi-attribute (MA), multi-relation (MR) and text rendering (TR); reasoning (R) covers logical (LR), behavioral (BR), hypothetical (HR), procedural (PR), generalization (GR), analogical (AR), commonsense (CR) and reconstructive (RR). Group scores recover unresolved judgments directly from the judge's final word (\cref{sec:app-evaluation}), and the first row gives the overall score, which averages all checklist items. Columns as in \cref{tab:text-loop-benchmarks}: zero text loops (0), one early text loop (1) or tied depth (=) at each image depth. Bold marks the best of the three.}
\label{tab:corebench-categories}
\begin{tabular}{@{}l*{9}{r}@{}}
\toprule
& \multicolumn{3}{c}{4 image loops} & \multicolumn{3}{c}{8 image loops} & \multicolumn{3}{c}{16 image loops} \\
\cmidrule(lr){2-4}\cmidrule(lr){5-7}\cmidrule(l){8-10}
Text loops & 0 & 1 & = & 0 & 1 & = & 0 & 1 & = \\
\midrule
Overall & 45.6 & \textbf{49.2} & 45.7 & 45.4 & \textbf{49.3} & 45.5 & 45.5 & \textbf{49.3} & 45.5 \\
\midrule
C-MA & 71.6 & \textbf{78.1} & 75.2 & 72.3 & \textbf{77.8} & 75.4 & 72.2 & \textbf{78.4} & 75.3 \\
C-MI & 70.0 & \textbf{73.5} & 70.4 & 69.8 & \textbf{74.5} & 70.3 & 69.6 & \textbf{74.2} & 70.3 \\
C-MR & 57.0 & \textbf{61.3} & 57.4 & 57.9 & \textbf{61.3} & 57.0 & 57.2 & \textbf{61.4} & 56.5 \\
C-TR & 17.6 & \textbf{32.0} & 13.3 & 14.7 & \textbf{29.6} & 12.2 & 15.3 & \textbf{28.9} & 11.8 \\
R-AR & 63.9 & \textbf{64.4} & 62.1 & 64.1 & \textbf{66.3} & 62.5 & 64.4 & \textbf{66.3} & 61.2 \\
R-BR & 22.9 & \textbf{26.2} & 25.5 & 24.3 & \textbf{26.8} & 25.8 & 24.0 & \textbf{25.9} & 25.7 \\
R-CR & 41.2 & \textbf{43.7} & 41.4 & 39.4 & \textbf{43.2} & 41.3 & 39.8 & \textbf{43.5} & 41.9 \\
R-GR & 43.3 & \textbf{44.5} & 44.4 & 42.8 & \textbf{44.5} & 43.8 & 43.0 & \textbf{44.0} & 43.5 \\
R-HR & 38.0 & \textbf{41.3} & 38.9 & 38.1 & \textbf{41.6} & 38.9 & 38.5 & \textbf{41.5} & 38.7 \\
R-LR & 33.7 & \textbf{34.2} & 33.0 & 32.3 & \textbf{34.3} & 32.5 & 33.1 & \textbf{34.9} & 34.1 \\
R-PR & 65.0 & \textbf{68.1} & 64.9 & 65.5 & \textbf{68.1} & 64.6 & 66.1 & \textbf{68.4} & 64.3 \\
R-RR & 23.0 & \textbf{23.5} & 22.2 & \textbf{23.9} & 23.7 & 22.4 & 22.9 & \textbf{23.7} & 22.6 \\
\bottomrule
\end{tabular}
\end{table}

\FloatBarrier
\subsection{Frequency Analysis Across Loops}
\label{sec:app-frequency}

We compare final images at successive doublings of image loop count across all 553 GenEval prompts and four noise seeds per prompt. Each prompt--seed pair uses the same initial noise at $K_I\in\{1,2,4,8,16\}$, with one text loop, 100 integration steps, and guidance scale 6. Let $I_K$ be the final RGB image in $[0,1]$ at image depth $K$. For each pair we take the channel-averaged Fourier power of $I_{2K}-I_K$ and sum it in three radial frequency bands. The bands correspond to wavelengths above 32 pixels, from just above 8 through 32 pixels, and at most 8 pixels. Their sum equals the mean squared RGB change by Parseval's identity. We compute each band's share from the mean power across all 2,212 pairs and resample prompts with their four seeds together for 2,000 paired bootstrap draws.

\begin{figure}[!htbp]
\centering
\includegraphics[width=\linewidth]{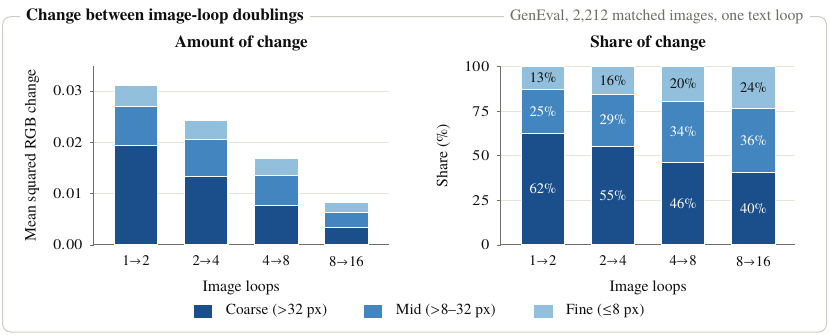}
\caption{\textbf{Later image-loop doublings make smaller changes, with a larger fine-scale share.} The left panel shows mean squared RGB change by spatial band. The right panel shows each band's share of that change. The analysis pairs 2,212 GenEval samples by prompt and initial noise.}
\label{fig:frequency-refinement}
\end{figure}

\Cref{fig:frequency-refinement} shows that mean squared RGB change falls from 0.0311 for one to two image loops to 0.0242, 0.0168, and 0.0084 for successive doublings. Change in the fine band also falls in absolute terms, from 0.0040 to 0.0020 across the same doublings. Coarse change falls faster, so wavelengths of at most 8 pixels account for 12.9\%, 15.5\%, 19.6\%, and 23.6\% of the remaining change. The difference between the first and last shares is 10.7 percentage points (95\% CI $[9.8,11.7]$). \Cref{sec:visual-loops} shows examples of these changes.

\FloatBarrier
\subsection{Caption Results}
\label{sec:caption-results}

\begin{figure}[!htbp]
\centering
\includegraphics[width=\linewidth]{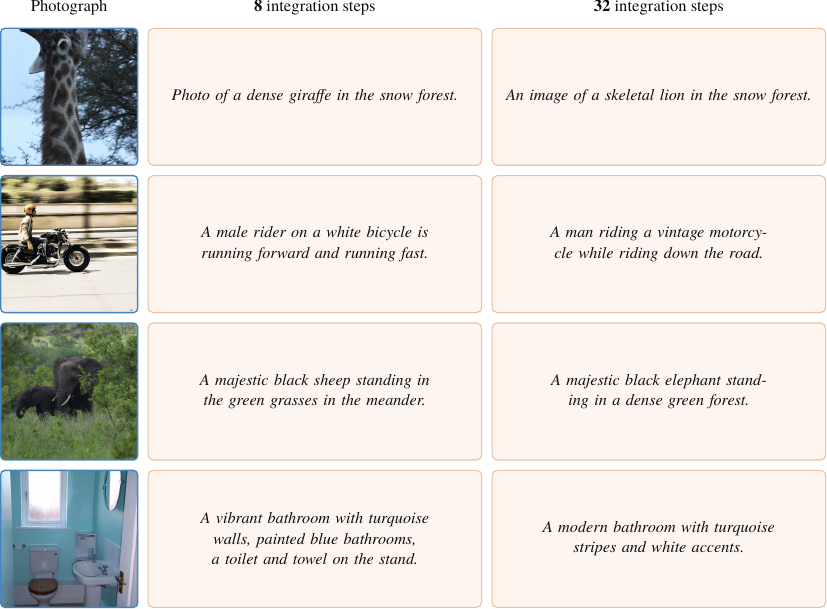}
\caption{\textbf{Eight and 32 integration steps get different subjects right.} At eight steps the caption names the giraffe and lists more of the bathroom, and at 32 steps it names the motorcycle and the elephant. Each row keeps its COCO photograph and initial noise, with zero image loops, eight text loops and guidance scale 2.}
\label{fig:caption-step-examples}
\end{figure}

These results give the caption scores behind \cref{sec:q1-depth,sec:q3-budget}, on the same 5,000 COCO photographs and scored with Long-CLIP as described in \cref{sec:app-evaluation}. \Cref{tab:caption-default-grid} gives the allocation grid at eight integration steps and guidance scale 6. \Cref{tab:caption-budget-guidance,tab:caption-budget-guidance-steps} sweep guidance at 32 and eight steps, \cref{tab:caption-budget-steps} sweeps integration steps, with examples in \cref{fig:caption-step-examples}, and \cref{tab:caption-budget-frontier} lists the settings on the caption frontier of \cref{fig:cc12m-compute}.

\begin{table}[!htbp]
\centering\small
\caption{\textbf{At eight integration steps and guidance scale 6, Long-CLIP changes little beyond two text loops or with image loops.} The same 5,000 COCO photographs and seeds as \cref{fig:cc12m-text}. Rows give image loops and columns text loops. H100 latency includes the frozen decoder and is the batch-one median of 20 timings after three warm-up runs.}
\label{tab:caption-default-grid}
\begin{tabular}{@{}lrrrrr@{}}
\toprule
& \multicolumn{5}{c}{Text loops $K_T$} \\
\cmidrule(l){2-6}
Image loops $K_I$ & 1 & 2 & 4 & 8 & 16 \\
\midrule
\multicolumn{6}{@{}l}{\emph{Long-CLIP}} \\
0 & 0.6753 & 0.6770 & 0.6781 & 0.6773 & 0.6771 \\
1 & 0.6758 & 0.6772 & 0.6775 & 0.6779 & 0.6777 \\
2 & 0.6756 & 0.6777 & 0.6777 & 0.6775 & 0.6779 \\
4 & 0.6752 & 0.6776 & 0.6775 & 0.6773 & 0.6777 \\
8 & 0.6754 & 0.6771 & 0.6775 & 0.6772 & 0.6777 \\
16 & 0.6753 & 0.6774 & 0.6771 & 0.6773 & 0.6774 \\
\midrule
\multicolumn{6}{@{}l}{\emph{Median seconds per caption}} \\
0 & 1.36 & 1.51 & 1.76 & 2.25 & 3.18 \\
1 & 1.49 & 1.62 & 1.86 & 2.34 & 3.30 \\
2 & 1.53 & 1.57 & 1.80 & 2.32 & 3.34 \\
4 & 1.79 & 1.83 & 1.96 & 2.40 & 3.41 \\
8 & 2.33 & 2.28 & 2.40 & 2.55 & 3.52 \\
16 & 3.33 & 3.36 & 3.44 & 3.54 & 3.86 \\
\bottomrule
\end{tabular}
\end{table}

\begin{table}[!htbp]
\centering\small
\setlength{\tabcolsep}{3.5pt}
\caption{\textbf{Caption guidance sweep on COCO.} Long-CLIP over text loops and guidance at 32 integration steps. Scale 1 disables guidance. Each setting scores the same 5,000 photographs.}
\label{tab:caption-budget-guidance}
\begin{tabular}{@{}lrrrrrrrrrr@{}}
\toprule
& \multicolumn{5}{c}{Text loops, 0 image loops} & \multicolumn{5}{c}{Text loops, 1 image loop} \\
\cmidrule(lr){2-6}\cmidrule(l){7-11}
Guidance & 1 & 2 & 4 & 8 & 16 & 1 & 2 & 4 & 8 & 16 \\
\midrule
1 & 0.6208 & 0.6284 & 0.6321 & 0.6340 & 0.6356 & 0.6226 & 0.6297 & 0.6334 & 0.6354 & 0.6364 \\
2 & 0.6460 & 0.6545 & 0.6585 & 0.6597 & 0.6602 & 0.6489 & 0.6566 & 0.6596 & 0.6604 & 0.6606 \\
3 & 0.6561 & 0.6643 & 0.6671 & 0.6689 & 0.6693 & 0.6582 & 0.6656 & 0.6677 & 0.6692 & 0.6693 \\
4 & 0.6604 & 0.6671 & 0.6707 & 0.6720 & 0.6730 & 0.6629 & 0.6694 & 0.6720 & 0.6731 & 0.6734 \\
5 & 0.6611 & 0.6692 & 0.6720 & 0.6731 & 0.6736 & 0.6643 & 0.6705 & 0.6732 & 0.6742 & 0.6748 \\
6 & 0.6622 & 0.6693 & 0.6719 & 0.6732 & 0.6743 & 0.6646 & 0.6709 & 0.6735 & 0.6745 & 0.6746 \\
\bottomrule
\end{tabular}
\end{table}

\begin{table}[!htbp]
\centering\small
\setlength{\tabcolsep}{3.5pt}
\caption{\textbf{At eight integration steps, Long-CLIP rises with guidance up to scale 6 and changes little beyond two text loops.} Long-CLIP on 5,000 COCO photographs with zero image loops. Scale 1 disables guidance. At scales 4 to 6, eight steps score highest with four text loops and 16 steps with sixteen. Dashes mark unevaluated settings.}
\label{tab:caption-budget-guidance-steps}
\begin{tabular}{@{}lrrrrrrrr@{}}
\toprule
& \multicolumn{5}{c}{Text loops, 8 steps} & \multicolumn{3}{c}{Text loops, 16 steps} \\
\cmidrule(lr){2-6}\cmidrule(l){7-9}
Guidance & 1 & 2 & 4 & 8 & 16 & 4 & 8 & 16 \\
\midrule
1 & 0.6307 & 0.6343 & 0.6356 & 0.6370 & 0.6378 & -- & -- & -- \\
2 & 0.6585 & 0.6614 & 0.6624 & 0.6623 & 0.6629 & 0.6609 & 0.6627 & 0.6628 \\
3 & 0.6680 & 0.6711 & 0.6710 & 0.6713 & 0.6706 & -- & -- & -- \\
4 & 0.6731 & 0.6751 & 0.6753 & 0.6748 & 0.6745 & 0.6745 & 0.6750 & 0.6755 \\
5 & 0.6747 & 0.6765 & 0.6772 & 0.6770 & 0.6770 & 0.6756 & 0.6769 & 0.6775 \\
6 & 0.6753 & 0.6770 & 0.6781 & 0.6773 & 0.6771 & 0.6773 & 0.6782 & 0.6787 \\
8 & -- & -- & 0.6769 & -- & -- & -- & -- & -- \\
\bottomrule
\end{tabular}
\end{table}

\begin{table}[!htbp]
\centering\small
\caption{\textbf{Caption integration-step sweep on COCO.} Long-CLIP at guidance scales 2 and 6. Dashes mark unevaluated settings.}
\label{tab:caption-budget-steps}
\begin{tabular}{@{}lrrrrr@{}}
\toprule
& \multicolumn{5}{c}{Integration steps} \\
\cmidrule(l){2-6}
$(K_I,K_T)$ & 4 & 8 & 16 & 32 & 64 \\
\midrule
\multicolumn{6}{@{}l}{\emph{Guidance scale 2}} \\
$(0,1)$ & -- & 0.6585 & -- & 0.6460 & -- \\
$(0,2)$ & -- & 0.6614 & -- & 0.6545 & -- \\
$(0,4)$ & 0.6530 & 0.6624 & 0.6609 & 0.6585 & 0.6570 \\
$(0,8)$ & 0.6529 & 0.6623 & 0.6627 & 0.6597 & 0.6581 \\
$(0,16)$ & 0.6530 & 0.6629 & 0.6628 & 0.6602 & 0.6596 \\
$(1,4)$ & -- & 0.6618 & 0.6615 & 0.6596 & -- \\
$(1,8)$ & -- & 0.6625 & 0.6625 & 0.6604 & -- \\
$(2,2)$ & -- & 0.6609 & 0.6597 & 0.6566 & -- \\
$(4,4)$ & -- & 0.6613 & 0.6619 & 0.6596 & -- \\
$(8,8)$ & -- & 0.6623 & 0.6627 & 0.6606 & -- \\
\midrule
\multicolumn{6}{@{}l}{\emph{Guidance scale 6}} \\
$(0,4)$ & 0.6669 & 0.6781 & 0.6773 & 0.6719 & 0.6666 \\
$(0,8)$ & 0.6659 & 0.6773 & 0.6782 & 0.6732 & 0.6678 \\
$(0,16)$ & 0.6659 & 0.6771 & 0.6787 & 0.6743 & 0.6684 \\
\bottomrule
\end{tabular}
\end{table}

\begin{table}[!htbp]
\centering\small
\caption{\textbf{Best observed caption scores within each compute budget.} The table lists each improvement in Long-CLIP among the 153 settings. Compute counts generator matrix multiplications per caption. H100 latency includes the frozen decoder and is the batch-one median of 20 timings after three warm-up runs. Dashes mark settings without a timing.}
\label{tab:caption-budget-frontier}
\begin{tabular}{@{}lrrrrrr@{}}
\toprule
$(K_I,K_T)$ & Guidance & Steps & TFLOPs & Long-CLIP & 95\% CI & Median s \\
\midrule
$(0,1)$ & 1 & 8 & 1.42 & 0.6307 & $[0.6294, 0.6321]$ & -- \\
$(0,2)$ & 1 & 8 & 1.76 & 0.6343 & $[0.6329, 0.6357]$ & -- \\
$(0,4)$ & 6 & 4 & 2.44 & 0.6669 & $[0.6652, 0.6685]$ & -- \\
$(0,1)$ & 6 & 8 & 2.84 & 0.6753 & $[0.6737, 0.6768]$ & 1.36 \\
$(0,2)$ & 6 & 8 & 3.52 & 0.6770 & $[0.6754, 0.6786]$ & 1.51 \\
$(0,4)$ & 6 & 8 & 4.88 & 0.6781 & $[0.6764, 0.6796]$ & 1.76 \\
$(0,8)$ & 6 & 16 & 15.23 & 0.6782 & $[0.6765, 0.6799]$ & 3.16 \\
$(0,16)$ & 6 & 16 & 26.16 & 0.6787 & $[0.6770, 0.6805]$ & 5.29 \\
\bottomrule
\end{tabular}
\end{table}

\begin{table}[!htbp]
\centering\small
\caption{\textbf{Captions depend on their photograph even without image loops.} We score each output against its original photograph after conditioning on the correct photograph, a shuffled photograph, or the learned null image.}
\label{tab:caption-conditioning-longclip}
\begin{tabular}{@{}lrrr@{}}
\toprule
$(K_I,K_T)$ & Condition & Long-CLIP & 95\% CI \\
\midrule
\multicolumn{4}{@{}l}{\emph{32 integration steps, guidance scale 2}} \\
$(0,8)$ & Normal & 0.6597 & $[0.6579, 0.6614]$ \\
$(0,8)$ & Shuffle & 0.5690 & $[0.5677, 0.5702]$ \\
$(0,8)$ & Null & 0.5709 & $[0.5698, 0.5719]$ \\
$(4,4)$ & Normal & 0.6596 & $[0.6579, 0.6613]$ \\
$(4,4)$ & Shuffle & 0.5693 & $[0.5681, 0.5705]$ \\
$(4,4)$ & Null & 0.5710 & $[0.5699, 0.5721]$ \\
\midrule
\multicolumn{4}{@{}l}{\emph{8 integration steps, guidance scale 6}} \\
$(0,8)$ & Normal & 0.6773 & $[0.6757, 0.6789]$ \\
$(0,8)$ & Shuffle & 0.5688 & $[0.5675, 0.5699]$ \\
$(0,8)$ & Null & 0.5747 & $[0.5738, 0.5757]$ \\
$(4,4)$ & Normal & 0.6775 & $[0.6758, 0.6791]$ \\
$(4,4)$ & Shuffle & 0.5694 & $[0.5681, 0.5706]$ \\
$(4,4)$ & Null & 0.5748 & $[0.5738, 0.5758]$ \\
\bottomrule
\end{tabular}
\end{table}

\paragraph{The photograph conditions the caption without image loops.}
The photograph enters through initial fusion before any loop. At zero image loops and eight text loops, conditioning each caption on a shuffled photograph lowers Long-CLIP, measured against the original photograph, from 0.6597 to 0.5690, and the learned null image gives 0.5709 (\cref{tab:caption-conditioning-longclip}). The training allocation $(4,4)$ drops by the same amount, to 0.5693 after shuffling. At eight steps and guidance scale 6, the drop is larger, from 0.6773 to 0.5688 at $(0,8)$.

\end{document}